\documentclass[twoside]{article}
\usepackage[round]{natbib}
\usepackage[utf8]{inputenc} 
\usepackage[T1]{fontenc}    
\usepackage[linktocpage,colorlinks,linkcolor=blue,anchorcolor=blue,citecolor=blue,urlcolor=blue,pagebackref]{hyperref}      
\renewcommand*{\backref}[1]{\ifx#1\relax \else Page #1 \fi}
\renewcommand*{\backrefalt}[4]{%
  \ifcase #1 \footnotesize{(Not cited.)}%
  \or        \footnotesize{(Cited on page~#2.)}%
  \else      \footnotesize{(Cited on pages~#2.)}%
  \fi
}
\usepackage{url}            
\usepackage{booktabs}       
\usepackage{amsfonts}       
\usepackage{nicefrac}       
\usepackage{microtype}      
\usepackage[dvipsnames]{xcolor}         
\usepackage{amsmath}
\usepackage{amsthm}

\usepackage{physics}
\usepackage{amsmath}
\usepackage{tikz}
\usepackage{mathdots}
\usepackage{yhmath}
\usepackage{cancel}
\usepackage{color}
\usepackage{siunitx}
\usepackage{array}
\usepackage{multirow}
\usepackage{amssymb}
\usepackage{gensymb}
\usepackage{tabularx}
\usepackage{extarrows}
\usepackage{booktabs}
\usetikzlibrary{fadings}
\usetikzlibrary{patterns}
\usetikzlibrary{shadows.blur}
\usetikzlibrary{shapes}
\usepackage{subcaption}

\usepackage{algorithm}
\usepackage{algpseudocode}
\newcommand{\hilight}[1]{\colorbox{lightgray}{#1}}

\newtheorem{theorem}{Theorem}[section]
\newtheorem{definition}[theorem]{Definition}
\newtheorem{assumption}[theorem]{Assumption}
\newtheorem{proposition}[theorem]{Proposition}
\newtheorem{corollary}[theorem]{Corollary}
\newtheorem{lemma}[theorem]{Lemma}

\newcommand{\kernel}{\kappa}
\newcommand{\DiffKernel}{v}

\newcommand*{\skipnumber}[2][1]{%
  {\renewcommand*{\alglinenumber}[1]{}\State #2}%
  \addtocounter{ALG@line}{-#1}}

\allowdisplaybreaks

\newcommand{\paperupdate}[1]{{#1}}

\usepackage{dsfont}
\newcommand{\mathbbm}[1]{\mathds{#1}}

\usepackage[accepted]{aistats2025}
\begin{document}

\runningtitle{Global Group Fairness in Federated Learning via Function Tracking}

\runningauthor{Yves Rychener, Daniel Kuhn,  Yifan Hu}

\twocolumn[

\aistatstitle{Global Group Fairness in Federated Learning via Function Tracking}

\aistatsauthor{ Yves Rychener${}^1$ \And Daniel Kuhn${}^1$ \And  Yifan Hu${}^{1,2}$ }

\aistatsaddress{${}^1$EPFL, Switzerland \And ${}^2$ETH Zurich, Switzerland }
]

\begin{abstract}
We investigate group fairness regularizers in federated learning, aiming to train a globally fair model in a distributed setting. Ensuring global fairness in distributed training presents unique challenges, as fairness regularizers typically involve probability metrics between distributions across all clients and are not naturally separable by client. To address this, we introduce a function-tracking scheme for the global fairness regularizer based on a Maximum Mean Discrepancy (MMD), which incurs a small communication overhead. This scheme seamlessly integrates into most federated learning algorithms while preserving rigorous convergence guarantees, as demonstrated in the context of FedAvg. Additionally, when enforcing differential privacy, the kernel-based MMD regularization enables straightforward analysis through a change of kernel, leveraging an intuitive interpretation of kernel convolution. Numerical experiments confirm our theoretical insights.
\end{abstract}

\section{INTRODUCTION} 
Many modern datasets are stored decentrally over multiple devices, organizations or jurisdictions and contain sensitive information. Federated learning~\citep{mcmahan2017communication} is proposed to train machine learning models on the aggregated (\emph{global}) dataset without exchanging data between local devices to protect privacy. On the other hand, these machine learning models are used in various aspects of daily life and can significantly impact people's lives. Unfortunately, cutting-edge AI models can learn or even amplify biases in datasets, which results in discriminatory predictions~\citep{ref:chouldechova2017fair,ref:dastin2018amazon,ref:propublica}. Ensuring that machine learning models treat different demographic groups fairly has become increasingly important. The desire to eliminate algorithmic bias has sparked the development of new algorithms in the field of fair machine learning. 
As both privacy and fairness become more critical, there is a need to train fair models on distributed datasets using federated learning. In this work, we study fair federated learning to protect privacy while eliminating algorithmic bias. 

\begin{figure}
    \centering
    \input{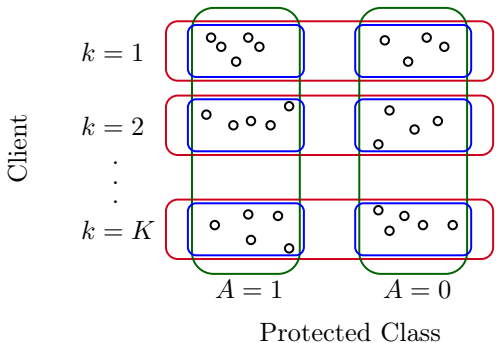}
    \caption{Illustration of client fairness (red), global group fairness (green) and local group fairness (blue)}
    \label{fig:fairness_illustration}
\end{figure}

In the following, we assume that each datapoint accommodates a binary attribute $A$ (called the sensitive attribute), which encodes membership in a demographic group we do not wish to discriminate against (e.g., race, gender, religion). 
We aim to train a model that satisfies a notion of group fairness on the global scale, that is, on the aggregated dataset (\textit{global group fairness}). We use Figure~\ref{fig:fairness_illustration} to explain how this fairness notion differs from others. To this end, let $\mathbb{P}_k$ be the distribution of the dataset of the $k$-th client, and consider the global mixture distribution $\mathbb{P}=\sum_{k=1}^K\nu_k\mathbb{P}_k$, where the mixture weight~$\nu_k$ encodes the size of the $k$-th dataset. In the following we say that a model $h$ is fair in the sense of demographic parity if 
the prediction $h(X)$ has the same distribution across several demographic groups. 

\textit{Global group fairness} holds if $\mathbb{P}^{h(X)|A=0}=\mathbb{P}^{h(X)|A=1}$, i.e., if the prediction distributions match across both demographic groups encircled in green in Figure~\ref{fig:fairness_illustration}.

\textit{Local group fairness} holds if $\mathbb{P}_k^{h(X)|A=0}=\mathbb{P}_k^{h(X)|A=1}$ for~all~$k$, i.e., if the prediction distributions match across all demographic groups and clients encircled~in~blue.

\emph{Client fairness} holds if $\mathbb{P}_k^{h(X)}=\mathbb{P}_{k'}^{h(X)}$ for all $k,k'$, i.e., if the prediction distributions match across all clients encircled in red.

By Simpson's paradox~\citep{simpson1951interpretation}, none of the above fairness notions implies the other. To see this, assume that $K=2$ and that $h(X)$ conditional on $A=a$ follows a shifted standard normal distribution with mean $0$ if $k+a$ is even and with mean $1$ if $k+a$ is odd. Thus, $h$ is globally fair because $\mathbb{P}^{h(X)|A=a}=\frac{1}{2}\mathcal{N}(0,1)+\frac{1}{2}\mathcal{N}(1,1)$ for all $a\in\{0,1\}$. However, $h$ does not satisfy local statistical parity because $\mathbb{P}_1^{h(X)|A=0}=\mathcal{N}(1,1)\neq\mathcal{N}(0,1)=\mathbb{P}_1^{h(X)|A=1}$. A similar situation will be explored in Section~\ref{sec:experiments:performance} below.

We emphasize that different applications necessitate different fairness criteria. For example, an algorithm that assesses the quality of images across multiple mobile phones may have to rate a similar percentage of images highly on each device in order to ensure user satisfaction (\emph{client fairness}). Similarly, multiple companies that train an algorithm for ranking job applicants may want to avoid racial discrimination at each individual company (\emph{local group fairness}). Finally,  multiple banks that train a model to predict their clients' credit worthiness may want to avoid gender discrimination across the entire population without exchanging proprietary data (\emph{global group fairness}). In practice, fairness criteria should be selected case-by-case. Global group fairness is not always the right choice.

Focusing on a general notion of group fairness, we propose a metric-regularized fair federated learning objective that incorporates a fairness-promoting regularization term into the classical federated learning framework. We show that a metric-based fairness regularizer cannot be decomposed as a weighted sum over local client objectives, making classical federated learning algorithms inapplicable. To address this, we introduce an approximation of the Maximum Mean Discrepancy (MMD) metric regularizer, updated at each communication round. This approach offers several advantages. First, the fairness-accuracy tradeoff can be easily adjusted by varying the regularization strength. Second, the metric-based regularization applies to a broad range of group fairness definitions. Third, it requires only a simple modification of existing federated learning methods. Finally, it is compatible with both cross-device and cross-silo settings and integrates seamlessly into most federated learning algorithms. We also demonstrate that differential privacy can be preserved while constructing the approximation.

Our main contributions are (1) a client-decomposable approximation of the metric-based fairness regularizer, (2) the derivation of the convergence rate for fairness-regularized FedAvg, and (3) an interpretation of differential privacy as a kernel change for MMD.

\paragraph{Notation}
We denote the space of probability distributions on $\mathcal{Z}\subseteq \mathbb{R}^d$ by $\mathcal{P}(\mathcal{Z})$. For any $\mathbb{P}\in\mathcal{P}(\mathcal{Z})$ and $E\subseteq\mathcal Z$, we denote the probability distribution of $Z$ conditioned on the event $Z\in E$ by $\mathbb{P}[\cdot|Z\in E]$. We sometimes use $\mathbb{P}^{Z|Z\in E}$ as shorthand for this distribution. We use subscripts to index different distributions (e.g., $\mathbb{P}_k$) and the respective expectation operators (e.g., $\mathbb{E}_k[\cdot]$). We further use $\mathcal{L}(\mathcal{Z}, \mathcal{Z}')$ to denote the space of Borel measurable functions from $\mathcal{Z}$ to $\mathcal{Z}'$ and, for a given $w\in\mathcal{L}(\mathcal{Z}, [0,\infty))$, we use $\mathcal{P}_w(\mathcal{Z})$ to denote the set of all $\mathbb{P}\in\mathcal{P}(\mathcal{Z})$ that satisfy $\int_\mathcal{Z}w(z)\mathbb{P}({\rm d}z)<\infty$.

\section{RELATED WORK}
\paragraph{Group Fairness} The seminal work of \citet{zafar2017fairness} introduced a covariance-based fairness constraint for logistic regression. Several studies extend this by proposing new fairness constraints \citep{ref:donini2018empirical, ref:wu2019convexity, ref:zafar2019fairness, ref:zafar2017fairness}. Rather than enforcing strict constraints, researchers often use fairness regularization for penalizing unfair models in the objective \citep{ref:donini2018empirical, ref:oneto2020expoliting, ref:taskesen2020distributionally, ref:wu2019convexity, ref:zafar2017fairness}, which enables the use of gradient-based methods for training deep neural networks. \cite{ref:lohaus2020} examine how fairness-regularized problems align with the original fairness constraint. A flexible and effective way to construct fairness regularizers is through metrics or divergences of probability distributions \citep{ref:oneto2020expoliting, prost2019toward, rychener2022metrizing, risser2022tackling}. For example, statistical parity can be enforced by penalizing differences between $\mathbb{P}^{h(X)|A=a}$ for $a\in\{0,1\}$. We refer to \citep{rychener2022metrizing} for discussions on fairness notions that can be addressed with this approach. In this paper, we use the MMD metric for fairness regularization due to its favorable computational properties. MMD metrics have already been used in centralized learning \citep{ref:oneto2020expoliting, prost2019toward, rychener2022metrizing}.

\paragraph{Federated Learning}
\cite{mcmahan2017communication} introduced the FedAVG algorithm, which trains models on distributed data while periodically communicating updates to a central node, enhancing data privacy. Since then, federated learning has garnered significant attention, leading to novel algorithms \citep{karimireddy2020scaffold, li2020federated, wang2019matcha}, convergence rate analyses \citep{karimireddy2020scaffold, khaled2020tighter, li2020federated, yu2019parallel}, and methods for enforcing differential privacy (DP) \citep{mcmahan2018learning, nikolaenko2013privacy, dwork2006calibrating} or leveraging homomorphic encryption \citep{nikolaenko2013privacy}. For a comprehensive overview of recent advances and open challenges, see~\citep{kairouz2021advances}.

\paragraph{Fair Federated Learning}
The federated learning literature considers three main notions of fairness. Client fairness arises in cross-device federated learning and ensures that individual clients are not disadvantaged \citep{mohri2019agnostic, deng2020distributionally, yue2021gifair}. Global group fairness more closely aligns with classical centralized group fairness, aiming to prevent systemic discrimination of individuals, ensuring similar treatment across demographic groups spanning multiple clients. One approach to achieving global group fairness in federated learning is adjusting FedAvg’s aggregation weights to promote a fairer aggregated model \citep{du2021fairness, ezzeldin2023fairfed, salazar2023fair}. Another direction extends centralized fairness algorithms to the federated setting, such as FairBatch \citep{roh2020fairbatch, zeng2021improving}, which adaptively reweights samples during local batch construction, $\min$-$\max$ fairness, which minimizes loss for the worst-off protected group \citep{papadaki2022minimax}, and pre- and post-processing techniques \citep{pentyala2022privfairfl}.

Some group fairness notions, particularly those defined in terms of expectations—where fairness is enforced by matching expectations rather then distributions across protected groups—are inherently client-decomposable. This allows existing federated learning algorithms to be applied directly \citep{chu2021fedfair, hu2024fair}. However, while useful in certain contexts, these fairness notions may be too simplistic in others. For instance, two prediction distributions with the same expectation but different variances would be considered fair under expectation-based fairness yet may still be unfair from a distributional perspective.

Lastly, local group fairness is studied by \cite{abay2020mitigating, cui2021addressing}. While beneficial in some cases, our Proposition~\ref{proposition:decomposition-impossibility} and recent findings by \cite{hamman2023demystifying} indicate that it does not necessarily ensure group fairness in the aggregated distribution. To our best knowledge, global group fairness regularizers have not yet been explored in federated learning.

\section{FAIR FEDERATED LEARNING WITH MMD METRICS}\label{sec:fair-FL}
We consider learning problems of the form
\begin{equation}\label{eq:learning-problem}
    \min_{h\in\mathcal{H}}\mathbb{E}[L(h(X), Y)],
\end{equation}
where $X\in\mathcal{X}\subseteq\mathbb{R}^d$ and $Y\in\mathcal{Y}\subseteq\mathbb{R}$ denote the feature vector and target, respectively, $L:\mathcal Y\times\mathcal Y\rightarrow\mathbb R$ stands for the loss function, and $h:\mathcal{X}\rightarrow\mathcal{Y}$ denotes a model chosen from a space $\mathcal{H}$ of Borel measurable functions. We further assume that there is a protected attribute $A\in\mathcal{A}=\{0,1\}$, which encodes a sensitive property of a person such as gender, race, or religion. Extending this framework to multiple or categorical protected attributes is feasible. The method to be developed below can accommodate such cases by incorporating additional fairness penalties into the objective.

We consider a federated learning setting with $K$ clients and represent datasets using probability distributions. While various data partitioning schemes have been explored in the federated learning literature (see~\citep{kairouz2021advances} for an overview), we focus on horizontal federated learning, where clients own different samples, all of which share the same feature space.  We assume that the global distribution $\mathbb{P} \in \mathcal{P}(\mathcal{X} \times \mathcal{Y} \times \mathcal A)$ of the features, targets and protected attributes can be represented as a mixture $\mathbb{P} = \sum_{k=1}^K \nu_k \mathbb{P}_k$ of local (client-wise) distributions $\mathbb{P}_k \in \mathcal{P}(\mathcal{X} \times \mathcal{Y} \times \mathcal A)$, with client weights $\nu_k \geq 0$ satisfying $\sum_{k=1}^K \nu_k = 1$.

To avoid discriminatory outcomes when solving~\eqref{eq:learning-problem}, we enforce a form of group fairness, which ensures that the demographic groups corresponding to $A=0$ and $A=1$ are treated similarly. 
We say that a hypothesis $h$ is $\varepsilon$-group fair~\citep{rychener2022metrizing} for some $\varepsilon\geq 0$~if
\begin{align*}
    & \mathcal{D}\left(\mathbb{P}^{S(h(X),Y))| A=0, (X,Y)\in \mathcal C_j}, \mathbb{P}^{S(h(X),Y)|A = 1, (X,Y)\in   \mathcal C_j}\right)\\
    & \hspace{2cm} \leq \varepsilon \quad \forall j\in\mathcal J,
\end{align*}
where $\mathcal{D}$ is a dissimilarity measure for probability distributions, $S:\mathcal{Y}\times\mathcal{Y}\rightarrow\mathbb{R}$ is a scoring function that assigns each prediction~$h(X)$ and output~$Y$ a score such as the task loss or some prediction score, and $\mathcal{C}=\{\mathcal{C}_j\}_{j\in\mathcal{J}}$ is a family of conditioning sets with $\mathcal{C}_j \subseteq \mathcal X\times\mathcal Y$. This generic fairness notion encompasses a variety of popular fairness criteria as shown in Table~\ref{tab:eps-fairness-illustration}.
For example, risk parity \citep{ref:donini2018empirical, ref:maity2021does} aims to match the distributions of the prediction loss $L(h(X),Y)$ across the two demographic groups, while equal opportunity \citep{ref:hardt2016equality, ref:pleiss2017fairness} aims to match the distributions of the prediction $h(X)$ across the subclasses of positively labeled datapoints within the two groups. In general, the {\em un}fairness of a hypothesis~$h$ is small if, for every~$j$, the distributions of the scores $S(h(X),Y)$ conditional on $A=a$ and $(X,Y)\in \mathcal{C}_j$ are similar across all $a\in\mathcal A$.  

For ease of exposition, from now on we focus on statistical parity, that is, we set $S(\widehat{y},y)=\widehat{y}$ and $\mathcal{C}=\{\mathcal{X}\times\mathcal{Y}\}$. However, all proposed methods readily extend to other group-fairness notions, including equal opportunity, equalized odds, and risk parity, see Appendix~\ref{app:other-notions}.
\begin{table}[t]
    \centering
    \caption{Common group fairness criteria induced by different scoring functions and families of conditioning sets (here, $\mathcal{X}_s$ is any conditioning set contained in $\mathcal{X}$)}
    \label{tab:eps-fairness-illustration}
    \resizebox{\linewidth}{!}{
    \begin{tabular}{l|cc}
Fairness Criterion & $S(\hat y, y)$  & $\mathcal{C}$\\\hline
Statistical parity  & \multirow{2}{*}{$\hat y$}      & \multirow{2}{*}{$\{\mathcal X \times \mathcal Y\}$} \\
 \citep{ref:agarwal2019fair}& &  \\ \hline
Equal opportunity  & \multirow{2}{*}{$\hat y$}      & \multirow{2}{*}{$\{\mathcal{X} \times \{1\}\}$} \\
 \citep{ref:hardt2016equality,ref:pleiss2017fairness} & & \\ \hline
Equalized odds & \multirow{2}{*}{$\hat y$}      & \multirow{2}{*}{$\{\mathcal{X} \times \{y\}\}_{y\in\mathcal Y}$} \\
\citep{ref:hardt2016equality, ref:pleiss2017fairness}& &  \\ \hline
Risk Parity & \multirow{2}{*}{$L(\hat y,y)$} & \multirow{2}{*}{$\{\mathcal X \times \mathcal Y\}$} \\
  \citep{ref:donini2018empirical, ref:maity2021does} & & \\ \hline
Conditional statistical parity & \multirow{2}{*}{$\hat y$} & \multirow{2}{*}{$\{\mathcal X_s \times \mathcal Y\}$} \\
\citep{corbett2017algorithmic} & & \\ \hline
Prob. predictive equality & \multirow{2}{*}{$\hat y$}      & \multirow{2}{*}{$\{\mathcal{X} \times \{0\}\}$} \\
  \citep{corbett2017algorithmic, ref:pleiss2017fairness} & & \\ \hline
\end{tabular}}\vspace{0.5cm}
\end{table}

Following \cite{ref:donini2018empirical, ref:oneto2020expoliting,rychener2022metrizing, ref:taskesen2020distributionally, ref:wu2019convexity, ref:zafar2017fairness}, we enforce group fairness in~\eqref{eq:learning-problem} by solving the fairness-regularized problem
\begin{equation}\label{eq:fair-learning-problem}
    \min_{h\in\mathcal{H}}\mathbb{E}[L(h(X), Y)]+\lambda\rho\left(\mathcal{D}(\mathbb{P}^{h(X)|A=0},\mathbb{P}^{h(X)|A=1})\right),
\end{equation}
where
$\rho:\mathbb R_+\rightarrow\mathbb R_+$ is a non-decreasing regularization function. Since problem~\eqref{eq:learning-problem} is client-decomposable, it can be directly solved using a federated learning approach. In contrast, problem~\eqref{eq:fair-learning-problem} lacks client-wise decomposition because the fairness regularizer couples clients in a nontrivial way. This observation is formally stated in the following proposition.

\begin{proposition}[Impossibility of Client-Decomposition]\label{proposition:decomposition-impossibility}
Let~$\rho:\mathbb R_+\rightarrow\mathbb R_+$ be a regularization function with $\rho(v)=0$ if and only if $v=0$. Then, there exists no dissimilarity measure $\mathcal{D}:\mathcal{P}(\mathcal{Y})\times\mathcal{P}(\mathcal{Y})\rightarrow\mathbb{R}_+$ satisfying the identity of indiscernibles such that
\begin{equation}
\label{eq:decomposition}
\begin{aligned}
    &\rho \left(\mathcal{D} \left(\sum_{k=1}^K \nu_k\mathbb{P}_k^{h(X)|A=0}, \sum_{k=1}^K \nu_k\mathbb{P}_k^{h(X)|A=1} \right) \right)\\ & ~ =\sum_{k=1}^K \nu_k\rho \left(\mathcal{D}(\mathbb{P}_k^{h(X)|A=0},\mathbb{P}_k^{h(X)|A=1}) \right)
\end{aligned}
\end{equation}
for all $h\in\mathcal H$, $\nu_k\geq 0$ with $\sum_{k=1}^K \nu_k=1$ and probability measures $\mathbb{P}_k\in\mathcal P(\mathcal X\times\mathcal Y\times\mathcal A)$, $k=1,\ldots, K$.
\end{proposition}

If a decomposition of the unfairness penalty, as described in Proposition~\ref{proposition:decomposition-impossibility}, existed, standard federated learning algorithms could solve the fairness-regularized problem by replacing the original local gradient estimator for~\eqref{eq:learning-problem} with a fairness-regularized local gradient estimator for~\eqref{eq:fair-learning-problem}, leveraging the right-hand side of~\eqref{eq:decomposition}. However, since such a decomposition is impossible, fairness-regularized federated learning presents a unique challenge, as the non-decomposability of group fairness regularizers appears to be intrinsic. This aligns with recent findings that local fairness does not generally imply global fairness, and vice versa, from an information-theoretic perspective~\citep{hamman2023demystifying}. The key question, then, is how to efficiently solve problem~\eqref{eq:fair-learning-problem} within a federated learning framework.

From now on, we will set the dissimilarity measure~$\mathcal{D}$ to a maximum mean discrepancy (MMD).

\begin{definition}[Maximum Mean Discrepancy]\label{def:mmd}
If $\kernel \in\mathcal L(\mathbb{R}^n \times \mathbb{R}^n,\mathbb{R})$ is a positive definite kernel and $w\in\mathcal L(\mathbb{R}^n ,[1,\infty))$ satisfies $\sup_{z\in\mathbb{R}^n} \kernel(z,z')/w(z)<\infty$ for all~$z'\in\mathbb{R}^n$, then the maximum mean discrepancy between $\mathbb{P}_1, \mathbb{P}_2 \in \mathcal{P}_w(\mathbb{R}^n)$ relative to~$\kernel$ is given by
\begin{align*}
    \mathcal{D}_{\rm MMD}(\mathbb{P}_1, \mathbb{P}_2) =\bigg(&\int_{\mathbb{R}^n \times \mathbb{R}^n} \kernel(z,z')\, \mathbb{P}_1(\dd z) \mathbb{P}_1(\dd z') \\
    + &\int_{\mathbb{R}^n \times \mathbb{R}^n}  \kernel(z, z') \, \mathbb{P}_2(\dd z)\mathbb{P}_2(\dd z') \\
    - &2\int_{\mathbb{R}^n\times \mathbb{R}^n} \kernel(z, z')\,  \mathbb{P}_1(\dd z) \mathbb{P}_2(\dd z') \bigg)^\frac{1}{2}.
\end{align*}
\end{definition}

Definition~\ref{def:mmd} characterizes the MMD distance between multivariate distributions on~$\mathbb R^n$. However, we defined the score $S(\widehat{y},y)$ as a scalar, and thus our fairness regularizer only requires an MMD distance for univariate distributions. Extensions to vector-valued scores are possible using a similar construction. Among all possible fairness regularizers in~\eqref{eq:fair-learning-problem} obtained by combining some regularization fundion~$\rho$ with a dissimilarity measure~$\mathcal D$, the
squared MMD distance stands out for multiple reasons. First, for common kernels~$\kernel$ and finite-dimensional parametrizations $h_\theta$ of the hypotheses in~$\mathcal H$, the fairness regularizer $\mathcal{D}_{\rm MMD}(\mathbb{P}^{h_\theta(X)|A=0},\mathbb{P}^{h_\theta(X)|A=1})^2$ constitutes a smooth function of the parameter~$\theta$.  
Second, in Section~\ref{sec:algorithm}, we introduce a biased local gradient estimator for this fairness regularizer based on a simple function tracking scheme. This estimator remains unbiased at the communication round, when client updates are aggregated by the server, but becomes biased during local updates performed between communication rounds. Despite this bias, existing convergence results for stochastic gradient descent (SGD) in federated learning can be naturally extended to account for it. Furthermore, the effect of differential privacy noise can be interpreted as a kernel convolution (see Section~\ref{sec:analysis}), which provides a clearer understanding of its impact on learning. This interpretation also allows generalization bounds to be directly applied~\citep{rychener2022metrizing}. For a rigorous restatement of this result, see Proposition~\ref{prop:generalization-bound} in the appendix. 
From now on we refer to the following instance of~\eqref{eq:fair-learning-problem} with $\mathcal{D}=\mathcal{D}_{\rm MMD}$ and $\rho(z)=\lambda z^2$ for some $\lambda\geq 0$ as the MMD-fair federated learning problem.
\begin{equation}\label{eq:mmd-learning-problem}
    \min_{h\in\mathcal{H}}\mathbb{E}[L(h(X), Y)]+\lambda\mathcal{D}_{\rm MMD}(\mathbb{P}^{h(X)|A=0},\mathbb{P}^{h(X)|A=1})^2.
\end{equation}

\section{ALGORITHM DESIGN}\label{sec:algorithm}
We now propose an efficient function tracking scheme for the fairness-regularized federated learning problem~\eqref{eq:mmd-learning-problem}. To this end, we focus on a parameterized hypothesis class $\mathcal{H}=\{h_\theta\}_{\theta\in\Theta}$, which includes models such as linear predictors and neural networks, and design a federated gradient-based algorithm. The objective function of problem~\eqref{eq:mmd-learning-problem} is henceforth denoted~as
$$
\begin{aligned}
 F(\theta) = & \mathbb{E}[L(h_\theta (X), Y)] \\
& +\lambda\mathcal{D}_{\rm MMD}(\mathbb{P}^{h_\theta (X)|A=0},\mathbb{P}^{h_\theta(X)|A=1})^2.  
\end{aligned}
$$
Note that $F(\theta)$ consists of the task loss $\mathbb{E}[L(h_\theta (X), Y)]$, which can be naturally decomposed into a weighted average of local losses $\mathbb{E}_k[L(h_\theta(X),Y)]$, and the fairness loss $\mathcal{D}_{\rm MMD}(\mathbb{P}^{h_\theta(X)|A=0},\mathbb{P}^{h_\theta(X)|A=1})^2$ weighted by~$\lambda$. 
 
In the absence of a fairness regularizer, the classical federated learning algorithm FedAvg \citep{mcmahan2017communication} operates as follows. At each communication round~$t$, the central server sends the current global model~$h_{\theta^t}$ to a selected subset of clients. Each participating client then updates~$h_{\theta^t}$ locally by performing~$E$ steps of stochastic gradient descent (SGD) on its own loss function $\mathbb{E}_k [L(h_\theta(X),Y)]$, where the expectation is taken with respect to the local data distribution~$\mathbb P_k$ of client~$k$. The client transmits the updated local model~$h_{\theta_{E+1,k}^t}$ back to the central server. The server then aggregates the received local models by computing their weighted average, where the contribution of each client is proportional to the number~$\nu_k$ of data samples it holds. Specifically, the new global model~$h_{\theta^{t+1}}$ is obtained by weighting each client's update according to its dataset size. This process is repeated for multiple communication rounds until convergence.

Unfortunately, Proposition~\ref{proposition:decomposition-impossibility} shows that the fairness loss does not admit a client-wise decomposition. This poses a challenge in federated learning, where transferring client data to a central server for computing gradients of the fairness loss and updating the global model is infeasible due to privacy constraints. To overcome this, we seek to incorporate fairness-aware updates at the client level. Specifically, we construct a client-decomposable approximation of the gradient of the fairness loss that each client can integrate into its local updates during the second step of FedAvg.

For a fixed parameter~$\theta$, we sample a finite number of predictions from $\mathbb{P}^{h_{\theta}(X)|A=a}$ across all clients' datasets and collect them in the set~$\mathcal{Y}_a$ for any~$a\in\mathcal A$. Using these datasets, we then construct the auxiliary function
\begin{align*}
    f(\theta;\mathcal{Y}_0,\mathcal{Y}_1)=2\bigg(&\int_{\mathbb{R}^n} C(h_\theta(x);\mathcal{Y}_0,\mathcal{Y}_1)\mathbb{P}^{X|A=0}(\dd x) \\- &\int_{\mathbb{R}^n} C(h_\theta(x);\mathcal{Y}_0,\mathcal{Y}_1)\mathbb{P}^{X|A=1}(\dd x)\bigg),
\end{align*}
where
\begin{equation*}
    C(\hat y;\mathcal{Y}_0,\mathcal{Y}_1) =\frac{1}{|\mathcal{Y}_0|}\sum_{y\in\mathcal{Y}_0} \kernel(\hat y,y) - \frac{1}{|\mathcal{Y}_1|}\sum_{y\in\mathcal{Y}_1} \kernel(\hat y,y).
\end{equation*}
Then, the gradient of~$f$ with respect to~$\theta$ is an unbiased estimator for the gradient of the fairness loss.
\begin{lemma}[Unbiased Gradient Estimator]\label{lemma:mmd-grad}
    If the kernel~$\kernel$ is continuous and the model $h_\theta(X)$ is $\mathbb{P}$-almost surely continuously differentiable in~$\theta$, then we have
    \begin{align*}
        &\nabla_\theta \mathcal{D}_{\rm MMD}\left(\mathbb{P}^{h_\theta(X)|A=0},\mathbb{P}^{h_\theta(X)|A=1}\right)^2\\& \quad =\mathbb{E} \left[\nabla_\theta f(\theta;\mathcal{Y}_0,\mathcal{Y}_1)\right].
    \end{align*}
\end{lemma}

We emphasize that $\nabla_\theta f(\theta;\mathcal{Y}_0,\mathcal{Y}_1)$ stands for the partial gradient of $f(\theta;\mathcal{Y}_0,\mathcal{Y}_1)$ with respect to its first argument. Thus, the gradient ignores the dependence of~$\mathcal{Y}_0$ and~$\mathcal{Y}_1$ on~$\theta$. We also emphasize that $f(\theta;\mathcal{Y}_0,\mathcal{Y}_1)$ is unavailable in practice because the distributions $\mathbb P^{X|A=a}$ for $a\in\mathcal A$ are unknown. However, if all clients were to share their data, then $f(\theta;\mathcal{Y}_0,\mathcal{Y}_1)$ could be estimated by sampling from these distributions.
By construction, the function~$f$ admits the client decomposition 
\begin{equation*}
    f(\theta;\mathcal{Y}_0,\mathcal{Y}_1) = \sum_{k=1}^K\nu_kf_k(\theta;\mathcal{Y}_0,\mathcal{Y}_1),
\end{equation*}
where the individual client functions $f_k$ are given by
\begin{align*}
    f_k(\theta;\mathcal{Y}_0,\mathcal{Y}_1)&=2\bigg(\alpha_k^0\int_{\mathbb{R}^n} C(h_\theta(x);\mathcal{Y}_0,\mathcal{Y}_1)\mathbb{P}_k^{X|A=0}(\dd x) \\&- \alpha_k^1\int_{\mathbb{R}^n} C(h_\theta(x);\mathcal{Y}_0,\mathcal{Y}_1)\mathbb{P}_k^{X|A=1}(\dd x)\bigg),
\end{align*}
and where $\alpha_k^a = \mathbb{P}_k(A=a)/\mathbb{P}(A=a)$ corrects the mixture weight~$\nu_k$ after conditioning on $A=a$. Note again that $f_k(\theta;\mathcal{Y}_0,\mathcal{Y}_1)$ is unavailable because the distributions~$\mathbb P_k^{X|A=a}$ for $a\in\mathcal A$ are unknown. However, the $k$-th client can estimate $f_k(\theta;\mathcal{Y}_0,\mathcal{Y}_1)$ by using only privately owned samples from these distributions. The above client decomposition of~$f$ readily leads to an estimator for the overall loss that is both unbiased and admits a client decomposition, that is, we have
\begin{equation*}
    F(\theta)\! =\! \mathbb E\!\left[\sum_{k=1}^K\nu_k\Big(\mathbb{E}_k[L(h_\theta(X),Y)] + \lambda f_k(\theta;\mathcal{Y}_0,\mathcal{Y}_1)\Big) \right]\!.
\end{equation*}

Note that Lemma~\ref{lemma:mmd-grad} does not contradict the impossibility result in Proposition~\ref{proposition:decomposition-impossibility} because the prediction datasets $\mathcal{Y}_0$ and $\mathcal{Y}_1$ appearing in $f_k(\theta;\mathcal{Y}_0,\mathcal{Y}_1)$ are constructed from information that originates from {\em all} clients. Thus, in order to construct $f_k(\theta;\mathcal{Y}_0,\mathcal{Y}_1)$, the $k$-th client requires information from other clients. Nevertheless, sharing $\mathcal{Y}_0$ and $\mathcal{Y}_1$ does not require transmitting each client's original dataset but rather a few scalar predictions, thereby preserving data privacy.

We now detail our proposed algorithm for fair federated learning; see also Algorithm~\ref{algo:fairfl}. It is inspired by the classical FedAvg algorithm and works as follows. At each communication round~$t$ and for each~$a \in \mathcal{A}$, the current model~$h_{\theta^t}$ is used to sample a finite number of predictions from $\mathbb{P}^{h_{\theta^t}(X) \mid A=a}$, which are collected in $\mathcal{Y}_a^t$. The prediction sets $\mathcal{Y}_0^t$ and $\mathcal{Y}_1^t$, along with the model~$h_{\theta^t}$, are transmitted to a randomly selected batch of clients. Each client then constructs unbiased gradient estimators for $\mathbb{E}_k[L(h_\theta(X),Y)] + \lambda f_k(\theta; \mathcal{Y}_0^t, \mathcal{Y}_1^t)$, performs multiple local SGD updates, and communicates the final iterate to the central server. The server subsequently aggregates the locally updated models to obtain $\theta^{t+1}$. The communication overhead of this algorithm exceeds that of standard FedAvg. Specifically, in each communication round~$t$, the server must exchange information about the sets $\mathcal{Y}_0^t$ and $\mathcal{Y}_1^t$ with the clients. However, the additional cost is moderate because these sets contain only one-dimensional samples. Indeed, the total bit size of $\mathcal{Y}_0^t$ and $\mathcal{Y}_1^t$ is often negligible compared to that of the high-dimensional iterate~$\theta^t$.

Lemma~\ref{lemma:mmd-grad} implies that, in each communication round~$t$, $\nabla_\theta f(\theta, \mathcal{Y}_0^t, \mathcal{Y}_1^t)$ serves as an unbiased estimator for $\nabla_\theta \mathcal{D}_{\rm MMD}(\mathbb{P}^{h_{\theta}(X) \mid A=0}, \mathbb{P}^{h_{\theta}(X) \mid A=1})^2$ at $\theta = \theta^t$. However, we emphasize that $\nabla_\theta f(\theta, \mathcal{Y}_0^t, \mathcal{Y}_1^t)$ is a biased gradient estimator for the fairness loss when evaluated at $\theta \neq \theta^t$, since $\mathcal{Y}_0^t$ and $\mathcal{Y}_1^t$ are generated under $\theta^t$. As these sets remain fixed throughout each communication round, the clients' SGD updates are biased. This bias is mitigated, however, as $\mathcal{Y}_0$ and $\mathcal{Y}_1$ are refreshed at the start of each communication round. Indeed, resampling prevents bias accumulation during local updates and ensures tracking of the population-level counterparts of the $C$~functions. The sizes of $\mathcal{Y}_0$ and $\mathcal{Y}_1$ affect the variances of the underlying estimators. However, as each client performs local SGD updates, the gradient estimator for the task loss also introduces variance. In theory, $O(1)$ samples suffice to construct $\mathcal{Y}_0$ and $\mathcal{Y}_1$.

The steps where Algorithm~\ref{algo:fairfl} differs from FedAvg are highlighted by gray boxes. On lines~2--3, an additional one-time communication between the clients and the server is required to compute the weights $\alpha_k^a = \mathbb{P}_k(A=a)/\mathbb{P}(A=a)$. In practice, these weights can be determined if the server knows the number of samples each client has from each demographic group~$a \in \mathcal{A}$. Exchanging sample size information is generally uncontroversial. On lines~5--6, $\mathcal{Y}_0^t$ and~$\mathcal{Y}_1^t$ must be generated and communicated to clients, incurring a moderate additional communication cost. As shown in Appendix~\ref{app:experiments:ablation}, using 50 samples for $\mathcal{Y}_0^t$ and $\mathcal{Y}_1^t$ suffices in our experiments. Furthermore, the potential information leakage from sharing these samples is minimal, especially since, as demonstrated in Section~\ref{sec:analysis:privacy}, they can be effectively protected with differential privacy. Finally, on line~11, the gradient of the fairness loss must be approximated in the client’s objective function. All these modifications are simple, compatible with most federated learning algorithms beyond FedAvg, and easy to implement.

\begin{algorithm}[!t]
\caption{MMD-Fair FedAvg}\label{algo:fairfl}
\begin{algorithmic}[1]
\State Initialize $\theta^1$
\State \hilight{Collect information to estimate $\alpha_k^a$, $a\in\mathcal A$}
\State \hilight{Communicate $\alpha_k^a$, $a\in\mathcal A$, to each client~$k$}
\For{$t = 1, \dots, T$}
    \State\hilight{Construct $\mathcal{Y}_a^t$, $a\in\mathcal A$, using $\theta^t$}
    \State Communicate $\theta^t$ \hilight{and $\mathcal{Y}_a^t$, $a\in\mathcal A$,} to all clients
    \State Sample a subset $\mathcal{S}\subseteq \{1,\ldots,K\}$ of $S$ clients
    \For{$k\in\mathcal{S}$ in parallel}
        \State Initialize $\theta_{1,k}^t = \theta^t$
        \For{local iteration $i = 1, \dots, E$}
            \State Compute a stochastic gradient $g_{i,k}^t$ with 
            \skipnumber{$\quad
            \mathbb{E} [g_{i,k}^t|\mathcal{Y}_0^t,\mathcal{Y}_1^t]$}
            \skipnumber{\quad \qquad$=\nabla_\theta \big(\mathbb{E}_k[L(h_{ \theta_{i,k}^t}(X),Y)]$}
            \skipnumber{\quad \qquad\qquad$ +\hilight{$ \lambda f_k( \theta_{i,k}^t;\mathcal{Y}_0^t,\mathcal{Y}_1^t)$}\big)
            $}
            \State Update $\theta_{i+1,k}^t = \theta_{i,k}^t - \eta_l g_{i,k}^t$
        \EndFor
        \State Communicate $\theta_{E+1,k}^t$ to server
    \EndFor
    \State Update $\theta^{t+1} = \theta^t + \eta_g \left(\frac{1}{S} \sum_{k\in\mathcal{S}} (\theta_{E+1,k}^t - \theta^t)\right)$
\EndFor
\end{algorithmic}
\end{algorithm}

\section{ANALYSIS OF ALGORITHM~\ref{algo:fairfl}}\label{sec:analysis}
We now analyze the convergence rate of Algorithm~\ref{algo:fairfl} and show how it can be made differentially private.

\subsection{Convergence Analysis}\label{sec:analysis:convergence}
\begin{theorem}[Convergence of Algorithm~\ref{algo:fairfl}]\label{thm:convergence-fairfl}
    Assume that the kernel function $\kernel$  and the loss function $L$ are Lipschitz continuous and Lipschitz smooth. Assume further that the model $h_\theta(X)$ is $\mathbb{P}$-almost surely Lipschitz continuous and Lipschitz smooth in $\theta$ and that the gradient estimator $g_{i,k}^t$ has bounded variance. Then, there exist a maximum stepsize~$\overline{\eta}>0$ and probabilities $\{w_t\}_{t=1}^T$ such that for any training hyperparameters $E\in\mathbb N$, $\eta_l>0$ and $\eta_g\geq1$ with $E\eta_l\eta_g\leq\overline{\eta}$, we have 
    \begin{align*}
        &\mathbb{E} \left\|\nabla_\theta F\left( \overline\theta \right) \right\|^2 \\ \leq &\mathcal{O}\left(\frac{1}{T+1}+\frac{1}{\sqrt{ES(T+1)}}+\frac{1}{E^{\frac{1}{3}}(T+1)^{\frac{2}{3}}}\right),
    \end{align*}
    where $\overline \theta = \theta^t$ with probability~$w_t$ and~$\theta^t$ is the $t$-th global iterate of Algorithm~\ref{algo:fairfl} for every~$t=1,\ldots, T$.
\end{theorem}

Our convergence analysis builds on that for FedAvg when the objective is nonconvex and only a random subset of clients participates in any round \citep{karimireddy2020scaffold}. It shows that Algorithm~\ref{algo:fairfl} requires the same number of communication rounds, local update steps, and clients per round as FedAvg. When clients are sampled randomly, applying momentum acceleration or variance reduction \citep{cheng2024momentum} can result in state-of-the-art convergence rates. However, as our focus is on integrating global fairness regularization into federated learning, we leave further convergence improvements using these ideas for future work. Our proof techniques are significantly different from those that are typically used in the context of biased gradient methods \citep{hu2024stochastic,hu2024multi}.

We briefly outline the proof idea for Theorem~\ref{thm:convergence-fairfl}. The analysis by~\cite{karimireddy2020scaffold} relates local update gradients to the gradient at the previous communication round. Even though the local gradient estimators are generally biased due to the fairness regularization, Lemma~\ref{lemma:mmd-grad} shows that they remain unbiased at the communication round when 
$\mathcal{Y}_0^t$ and $\mathcal{Y}_1^t$ are sampled. As a result, the bias from fairness regularization is reset at each communication round. We use this property to analyze the bias similarly to how \cite{karimireddy2020scaffold} analyze local gradients. The full proof is provided in Appendix~\ref{app:analysis:convergence}.

\subsection{Differential Privacy}
\label{sec:analysis:privacy}
To address privacy concerns arising from sharing predictions $h_\theta(X)$ and parameters $\theta_{E+1,k}^t$, federated learning commonly employs differential privacy (DP) to prevent sensitive information from being leaked through network communications. Specifically, additive DP mechanisms introduce random noise to shared parameters, ensuring that clients cannot reconstruct data samples from other participants. The protection of model updates $\theta_{E+1,k}^t$ has already been studied extensive in the federated learning literature (see~\citep{el2022differential} for a survey). We therefore show how DP can be used to safeguard shared predictions $h_\theta(X)$. This can be achieved by adding Gaussian noise (the {\em Gaussian mechanism} \citep{balle2018improving}) or Laplacian noise (the {\em Laplacian mechanism} \citep{koufogiannis2015optimality}) to the predictions shared on line~5 of Algorithm~\ref{algo:fairfl}. For each prediction $\widehat{Y} \in \mathcal{Y}_a$, $a\in \mathcal A$, we construct a privacy-protected version $\Tilde{Y} = \widehat{Y} + \xi$ with $\xi \sim \mu_{\rm DP}$, and we denote by $\Tilde{\mathcal{Y}}_a$ the resulting privacy-protected set. If the predictions $\widehat{Y}$ are bounded, as is the case for scores after a sigmoid activation in neural networks, the variance (for the Gaussian mechanism) or scale (for the Laplacian mechanism) can be calibrated and remains finite. Even in regression tasks, it is often reasonable to assume that $\widehat{Y}$ is bounded or can be clipped, as most real-world quantities to be estimated are naturally constrained. While DP noise is typically seen as a nuisance, we now show that noise added to shared predictions $\widehat{Y}$ can instead be interpreted as a kernel modification in the MMD metric.

\begin{proposition}[Effect of Differential Privacy]\label{proposition:DP-effect}
    Assume that $\xi\sim\mu_{\rm DP}$, $a,b\overset{i.i.d.}{\sim}\mu'$ and $\xi \overset{d}{=} a-b$. Construct the differentially private sets $\Tilde{\mathcal{Y}}_0$ and $\Tilde{\mathcal{Y}}_1$ by adding i.i.d.\ samples from~$\mu_{\rm DP}$ to each element of $\mathcal{Y}_0$ and $\mathcal{Y}_1$, respectively. If the kernel $\kernel$ satisfies $\kernel(x,y)=\DiffKernel(x-y)$ for some real-valued function $\DiffKernel$, then we have 
    \begin{align*}
        \Tilde{C}(z;\mathcal{Y}_0,\mathcal{Y}_1) & := \mathbb{E} \left[\left. C(z;\Tilde{\mathcal{Y}}_0,\Tilde{\mathcal{Y}}_1) \right| \mathcal{Y}_0,\mathcal{Y}_1 \right] \\&=\frac{1}{|\mathcal{Y}_0|}\sum_{y\in\mathcal{Y}_0}\Tilde{\kernel}(z,y)-\frac{1}{|\mathcal{Y}_1|}\sum_{y\in\mathcal{Y}_1}\Tilde{\kernel}(z,y),
    \end{align*}
    that is, 
    $\Tilde{C}(z;\mathcal{Y}_0,\mathcal{Y}_1)$ is a variant of $C(z;\mathcal{Y}_0,\mathcal{Y}_1)$, where the original kernel $\kappa(x,y)$ is replaced with the DP kernel $\Tilde{\kernel}(x,y)=\int_\mathbb{R} \kernel(x,y+\xi)\mu_{\rm DP}(\rm d\xi)$. Note that $\Tilde{\kernel}(x,y)$ can be viewed as the convolution of $\mu_{\rm DP}$ and $\kernel$. Furthermore, if $\kernel$ is positive semidefinite, then so is $\Tilde{\kernel}$.
\end{proposition}

The assumptions of Proposition~\ref{proposition:DP-effect} hold for many DP mechanisms and distributions of interest. The conditions on $\mu_{\rm DP}$ are satisfied by both the Gaussian mechanism (where $\mu'$ is a Gaussian distribution with lower variance than $\mu_{\rm DP}$) and the Laplacian mechanism (where $\mu'$ is an exponential distribution). The condition on $\kappa$ holds for many commonly used kernels, such as the Gaussian and Laplacian/Abelian kernels. However, one can show that the condition $\kernel(x,y)=v(x-y)$ implies $|\kernel(x,y)|\leq |v(0)|$, which can only be satisfied by bounded kernels. By Proposition~\ref{proposition:DP-effect}, adding differential privacy ensures that $f$ remains an unbiased gradient of the MMD distance between prediction distributions at the communication round, preserving the validity of all preceding arguments. However, the expected gradient corresponds to a different objective—namely, the MMD-fair learning problem~\eqref{eq:mmd-learning-problem} with kernel $\Tilde{\kernel}$ instead of $\kernel$. An interesting corollary of Proposition~\ref{proposition:DP-effect} arises when differential privacy is enforced using a Gaussian mechanism.

\begin{corollary}[Differentiable Privacy with Gaussian Kernel]\label{corr:DP-gaussian}
    Under the assumptions of Proposition~\ref{proposition:DP-effect}, if $\kernel$ is the Gaussian Kernel and $\mu_{\rm DP}$ is a Gaussian distribution (i.e., a Gaussian mechanism is used), then $\Tilde{\kernel}$ is a Gaussian kernel with increased variance.
\end{corollary}

Corollary~\ref{corr:DP-gaussian} immediately follows from Proposition~\ref{proposition:DP-effect} because the family of Gaussian distributions is closed under convolution.

\section{EXPERIMENTS}\label{sec:experiments:performance}
We conduct classification experiments on both synthetic and real data to demonstrate that MMD-regularized federated learning performs comparably to centralized training on the global dataset while surpassing other fair federated learning methods in terms of global statistical parity. In all experiments, we report the degree of SP unfairness of the tested classifiers, which is defined as the absolute difference between the positive prediction frequencies in the two demographic groups corresponding to~$A=0$ and~$A=1$, respectively. All results are averaged over 10 different random seeds, and we use a distance-based kernel for training.

\begin{figure}[t!]
    \centering
    \begin{subfigure}[b]{0.25\textwidth}
        \centering
        \includegraphics[trim={0.5cm 0.5cm 0.5cm 0.5cm},clip,width=0.9\linewidth]{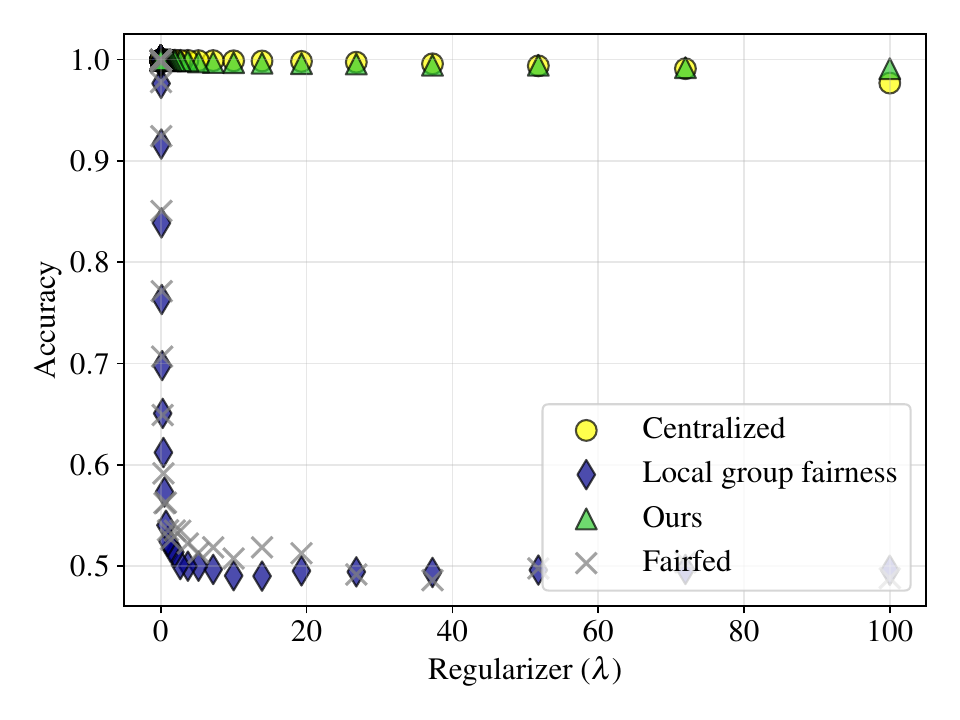}
        \caption{Accuracy Profile}
    \end{subfigure}%
    \begin{subfigure}[b]{0.25\textwidth}
        \centering
        \includegraphics[trim={0.5cm 0.5cm 0.5cm 0.5cm},clip,width=0.9\linewidth]{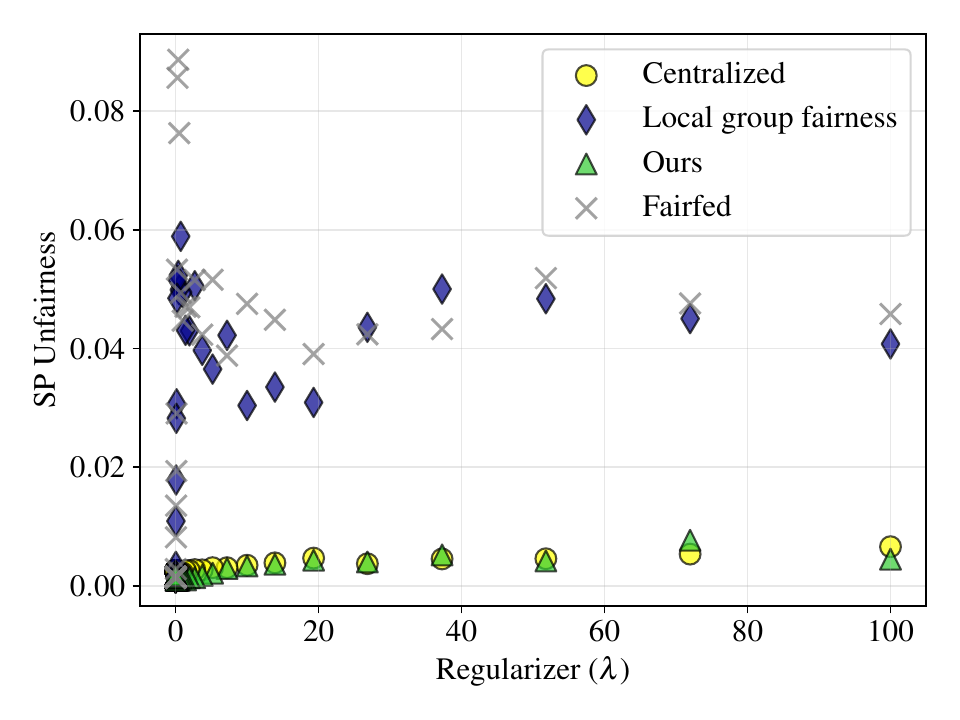}
        \caption{Unfairness Profile}
    \end{subfigure}
    \caption{Performance comparison on synthetic data}\label{fig:exp:synthetic}
\end{figure}

\paragraph{Synthetic Data}
First, to highlight the advantage of a global fairness regularizer, we illustrate the implications of Proposition~\ref{proposition:decomposition-impossibility} on learning performance. Consider a setting with $K=10$ clients, each with equal weight ($\nu_k=\frac{1}{10}$ for all $k=1,\ldots,K$). For all $k=1,\ldots,K$, we assume that $\mathbb{P}_k[A=0]=\mathbb{P}_k[A=1]=\frac{1}{2}$ and that $X\sim\mathcal{N}(\mu(k,A),\mathbb{I})$, where $\mathbb{I}$ is the $10$-dimensional identity matrix, and $\mu(k,a)$ is a vector with all entries equal to $1$ ($-1$) if $k+a$ is even (odd). The label is set to $Y=1$ if $\mathbbm{1}^\top X>0$ and to $Y=0$ otherwise. We sample 200 data points per client and compare our approach to two baselines. In the \emph{centralized} method, all clients send their data to a server, which solves~\eqref{eq:mmd-learning-problem} on the pooled dataset. In the \emph{local group fairness} method, federated learning is performed, but each client~$k$ uses the local fairness regularizer 
\begin{equation*}
    \mathcal{D}_{\rm MMD}\left(\mathbb{P}_k^{h_\theta(X)|A=0},\mathbb{P}_k^{h_\theta(X)|A=1}\right)^2.
\end{equation*}  
We also compare against FairFed~\citep{ezzeldin2023fairfed}, which employs an adaptive, fairness-aware client reweighting and can be combined with local fairness methods. For a fair comparison, we integrate FairFed’s global fair aggregation with MMD-based local fair training. All methods train a fairness-regularized logistic regression model. Figure~\ref{fig:exp:synthetic} presents accuracy and unfairness profiles for different regularization weights~$\lambda$. Our approach, like the centralized method, maintains accuracy even for higher values of $\lambda$, but it achieves global fairness already for low $\lambda$. In contrast, the local group fairness method significantly biases the model when fairness regularization is imposed, and reducing client-level unfairness does not necessarily lower global unfairness. FairFed suffers from similar problems, likely due to its combination of fair aggregation with a local fairness penalty. The Bayes classifier on the global distribution $\mathbb{P}$, given by $\mathbbm{1}[\mathbbm{1}^\top X>0]$, achieves perfect accuracy while satisfying statistical parity, as $\mathbb{P}[\mathbbm{1}^\top X>0|A=0]=\mathbb{P}[\mathbbm{1}^\top X>0|A=1]=\frac{1}{2}$. However, when restricted to a single client, the Bayes classifier no longer satisfies statistical parity.

\paragraph{Communities $\&$ Crime Dataset} 
We evaluate our algorithm on the Communities $\&$ Crime dataset.\footnote{\url{http://archive.ics.uci.edu/ml/datasets/Communities+and+Crime}} This dataset includes law enforcement and socio-economic data for 1,994 US communities, described by 99 features. The task is to classify whether the incidence of violent crimes exceeds the national average. Following~\cite{ref:calders2013controlling}, we define a binary protected attribute by thresholding the percentage of Black or African American residents at the median. We assume data is stored by state, with states unwilling to share their communities' data. States with fewer than four communities are removed. All methods train a two-layer feedforward neural network with 16 hidden units and ReLU activation. We compare our approach, which solves~\eqref{eq:mmd-learning-problem} using Algorithm~\ref{algo:fairfl} with $|\mathcal{Y}_0|=|\mathcal{Y}_1|=100$, against several baselines: a \emph{centralized} model trained on aggregated data using SGD on~\eqref{eq:mmd-learning-problem}; FedAvg with a local MMD-fairness regularizer (\emph{local group fairness}); \emph{agnostic} federated learning~\citep{mohri2019agnostic}, which enforces client fairness without additional hyperparameters; \emph{FedFB}, a federated version of FairBatch~\citep{zeng2021improving}; and FairFed~\citep{ezzeldin2023fairfed}, which combines fair aggregation with local fairness penalties. Hyperparameter details are provided in Appendix~\ref{app:hyperparams}. Figure~\ref{fig:experiments:CC} presents the Pareto frontier of the accuracy-unfairness tradeoff as we vary the regularization parameter $\lambda$. We observe that~\citep{mohri2019agnostic} fails to induce group fairness, as client fairness does not necessarily imply global group fairness. Furthermore, our method outperforms the local group fairness approach, indicating bias heterogeneity across clients, consistent with our synthetic experiment. Finally, FairFed performs comparably to our method, and both achieve a fairness-accuracy tradeoff similar to training on aggregated data.

\begin{figure}
    \centering
    \begin{subfigure}[b]{0.45\textwidth}
    \centering
    \includegraphics[height=0.4\linewidth]{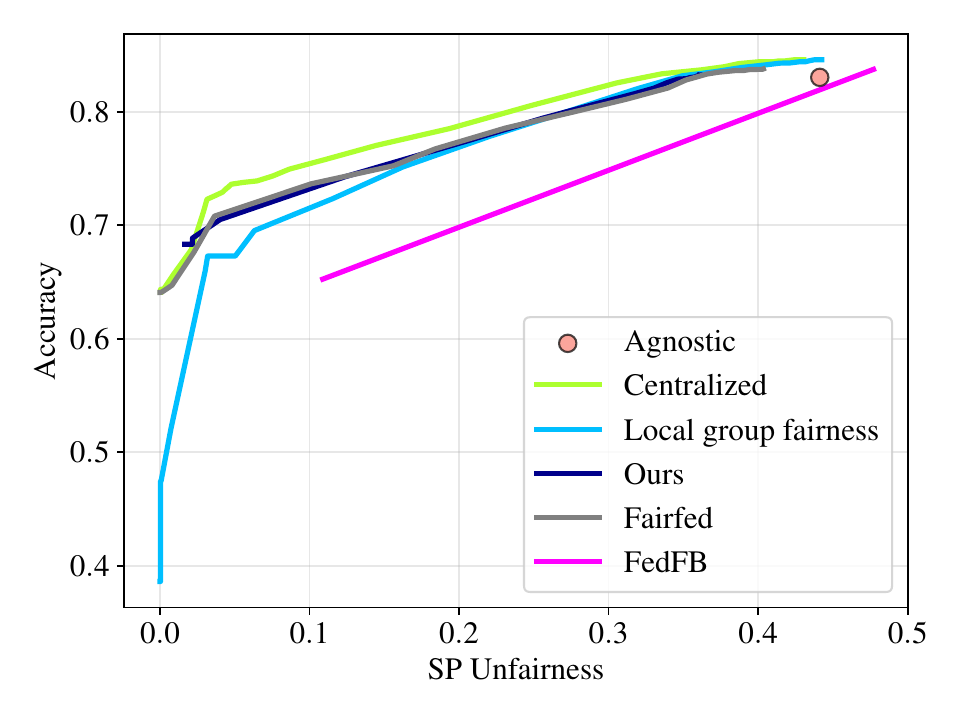}
    \caption{Communities$\&$Crime dataset}
    \label{fig:experiments:CC}
    \end{subfigure}
    \begin{subfigure}[b]{0.45\textwidth}
    \centering
    \includegraphics[height=0.4\linewidth]{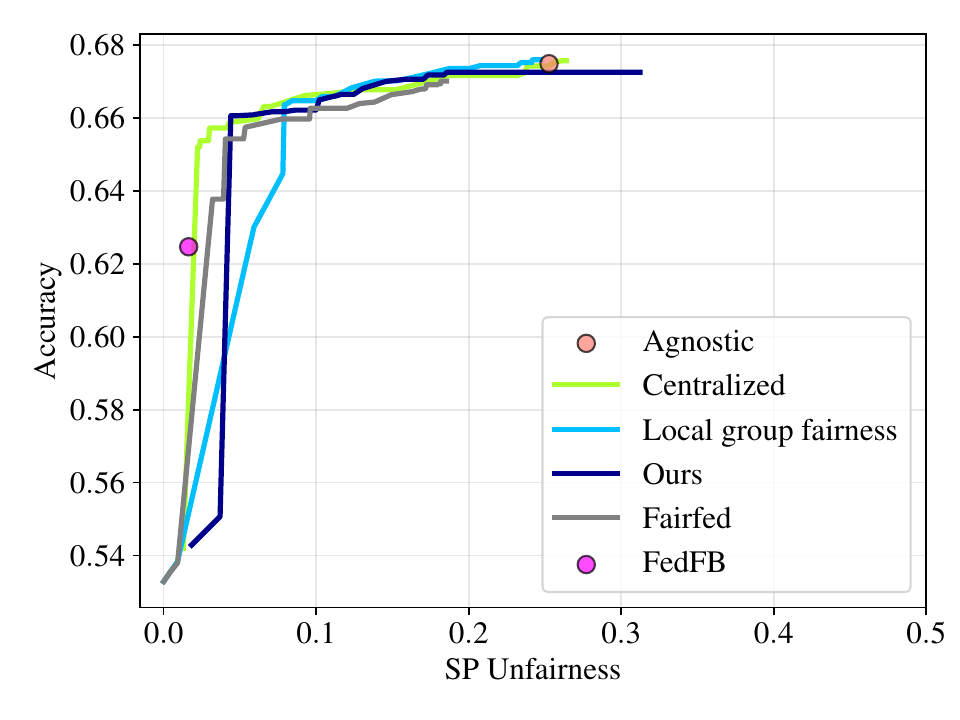}
    \caption{COMPAS dataset}
    \label{fig:experiments:compas}
    \end{subfigure}
    \caption{Performance comparison on real data}\label{fig:experiments}
\end{figure}

\paragraph{Compas Dataset} 
For our next experiment, we consider the COMPAS dataset.\footnote{\url{https://www.kaggle.com/danofer/compass}} It contains records of defendants in Broward County, Florida, described by eight features, including age, gender, race, and prior offenses—features used by the COMPAS (Correctional Offender Management Profiling for Alternative Sanctions) algorithm. We use race as the protected attribute and predict recidivism within two years of the original offense. After preprocessing, the dataset contains 5,278 datapoints, which we assume are distributed across $K=3$ clients, split by age group (25 or younger, 25–45, and 45 or older). All hyperparameters and baselines remain the same as in the Communities $\&$ Crime experiment. Figure~\ref{fig:experiments:compas} presents the Pareto frontier of the accuracy-unfairness tradeoff. At high accuracy levels, both our method and the local group fairness approach perform comparably to the centralized baseline. However, the local group fairness method sacrifices significant accuracy to achieve fairness below 0.1, whereas our approach maintains competitiveness with the centralized method even at fairness values of 0.05. For stronger regularization, our method exhibits lower accuracy and higher unfairness, which we attribute to noise from sampling the sets $\mathcal{Y}_0$ and $\mathcal{Y}_1$. Addressing this would require tuning the step size and the sizes of $\mathcal{Y}_0$ and $\mathcal{Y}_1$, but we opted against this to ensure a fair comparison. FedFB achieves strong fairness but at the cost of accuracy, and its accuracy-fairness tradeoff is not adjustable. FairFed performs well but does not match the performance of our method.  

In summary, the proposed fairness-regularized algorithm produces globally fair predictors without unnecessary accuracy loss. Moreover, it performs comparably to centralized training while sharing only $|\mathcal{Y}_0|=|\mathcal{Y}_1|=100$ samples. As shown in Appendix~\ref{app:experiments:ablation}, the algorithm is robust to variations in the sample~sizes.

\section{Conclusions}
We introduce a global fairness regularizer for federated learning and propose a distributed training method to enhance global group fairness in federated algorithms. When applying such regularizers, it is essential to carefully assess the appropriate fairness notion (if any) and recognize that fairness regularization can have unintended effects, particularly on heavily biased distributions. Our approach is compatible with most group fairness notions, as shown in Table~\ref{tab:eps-fairness-illustration}. Additionally, federated learning algorithms like FedAvg can lead to data leakage even when only gradients are shared. Thus, to strengthen security and privacy in practical applications, our methods should be combined with differential privacy \citep{dwork2006calibrating,mcmahan2018learning} or homomorphic encryption \citep{nikolaenko2013privacy}.

\paragraph{Acknowledgments}
This work was supported as a part of the NCCR Automation, a National Centre of Competence in Research, funded by the Swiss National Science Foundation (grant number 51NF40\_225155).

\bibliographystyle{plainnat} 
\bibliography{main}

\newpage
\onecolumn
\appendix
\section*{\LARGE Appendix}
\section{Auxiliary Results}
\subsection{Generalization Bound}
We formally restate that generalization bound by~\citet{rychener2022metrizing}. To this end, we denote the empirical distribution constructed from~$N$ training samples $\{(X_i,Y_i,A_i)\}_{i=1}^N$ by $\widehat{\mathbb{P}}$.
\begin{proposition}[Generalization Bounds {\citep[Theorem~A.2]{rychener2022metrizing}}]
\label{prop:generalization-bound}
    Suppose $\mathcal{H} \subseteq \mathcal L(\mathcal{X},\{0,1\})$ is a family of binary functions with VC dimension~$\mathcal{V}$, and let $\mathcal D_{\rm MMD}$ be the MMD distance induced by  a kernel~$\kernel$. Select any~$N\in\mathbb N$, and define $N_1= \sum_{i=1}^NA_i$. Then, for any~$h \in \mathcal{H}$ and $\delta\in(0,1)$, conditioned on the event $\mathcal{V}<N_1< N-\mathcal{V}$, the accuracy and MMD unfairness of~$h$ jointly satisfy
    \begin{align*}
        \mathbb{P}[h(X)=Y] &\geq \widehat{\mathbb{P}}[h(X)=Y]-\frac{1}{2}\sqrt{\frac{4}{N}\left(\mathcal{V}\left(\log\left(\frac{2N}{\mathcal{V}}\right)+1\right)-\log\left(\delta/12\right)\right)}
    \end{align*}
    and
    \begin{align*}
        \mathcal{D}_{\rm MMD} \left( \mathbb{P}^{h(X) | A = 0} ,  \mathbb{P}^{h(X) | A=1} \right) &\leq  \mathcal{D}_{\rm MMD}\left(\widehat{\mathbb{P}}^{h(X) | A = 0} , \widehat{\mathbb{P}}^{h(X) | A=1} \right) + \frac{\alpha}{2}\sqrt{\frac{4}{N_1}\left(\mathcal{V}\left(\log\left(\frac{2N_1}{\mathcal{V}}\right)+1\right)-\log(\delta/12)\right)}\\&+ \frac{\alpha}{2}\sqrt{\frac{4}{N-N_1}\left(\mathcal{V}\left(\log\left(\frac{2(N-N_1)}{\mathcal{V}}\right)+1\right)-\log(\delta/12)\right)}
    \end{align*}
    with probability at least $1-\delta$, where $\alpha=\sup_{\psi\in \Psi}\left|\psi(0)-\psi(1)\right|$ and $\Psi=\{\psi\in\mathbb{H}_\kernel:\|\psi\|_{\mathbb{H}_\kernel}\leq 1\}$ denotes the unit ball in the reproducing kernel Hilbert space generated by the kernel~$\kernel$.
\end{proposition}
We note that Proposition~\ref{prop:generalization-bound} readily extends to other fairness notions beyond statistical parity.

\paperupdate{
\subsection{Application to Other Fairness Notions}\label{app:other-notions}
We now briefly outline how Algorithm~\ref{algo:fairfl} can be extended to other group fairness notions, such as those in Table~\ref{tab:eps-fairness-illustration}.  
First, we require a fairness regularizer for each conditioning set $\mathcal{C}_j \in \mathcal{C}$. Indexing the regularizers by $j \in \mathcal{J}$, the following changes must be made compared to Algorithm~\ref{algo:fairfl} described in Section~\ref{sec:algorithm}.

\begin{enumerate}
    \item Instead of sampling predictions from $\mathbb{P}^{h_\theta(X)| A=a}$ and collecting them in $\mathcal Y_a$ for any $a\in\mathcal A$, we must sample predictions from $\mathbb{P}^{S(h_{\theta}(X),Y)| A=a,\, (X,Y) \in \mathcal{C}_j}$ and collect them in $\mathcal{Y}_{a,j}$ for every $a\in\mathcal A$ and $j\in\mathcal J$.
    
    \item Instead of decomposing~$f$ into client-wise functions~$f_k$, we need to decompose an analogue~$f^j$ of~$f$, one for every $j\in\mathcal J$, into client-wise functions~$f_k^j$. These functions are defined as
        \begin{align*}
            f_k^j(\theta;\mathcal{Y}_{0,j},\mathcal{Y}_{1,j})=2\Big(&\alpha_{k,j}^0\int_{\mathbb{R}^n} C\Big(S(h_{\theta^t}(X),Y);\mathcal{Y}_{0,j},\mathcal{Y}_{1,j}\Big)\mathbb{P}_k^{X|A=0,\, (X,Y) \in \mathcal{C}_j}(\dd x) \\-&\alpha_{k,j}^1\int_{\mathbb{R}^n} C \Big(S(h_{\theta^t}(X),Y);\mathcal{Y}_{0,j},\mathcal{Y}_{1,j}\Big)\mathbb{P}_k^{X|A=1,\, (X,Y) \in \mathcal{C}_j}(\dd x)\Big).
        \end{align*}
    The main differences to the definition of~$f_k$ are that we evaluate~$C$ on $S(h_{\theta^t}(X),Y)$ instead of $h_{\theta^t}(X)$ and that we not only condition on $A=a$ but also on $(X,Y)\in\mathcal C_j$.
    
    \item Instead of computing the weights~$\alpha^a_k$, we need to construct more structured weights $\alpha^a_{k,j}$ by not only conditioning on on $A=a$ but also on $(X,Y)\in\mathcal C_j$. Specifically, we define
    \begin{align*}
        \alpha_{k,j}^a &= \frac{\mathbb{P}_k[A=a ,\,(X,Y)\in\mathcal{C}_j]}{\mathbb{P}[A=a,\, (X,Y)\in\mathcal{C}_j]}.
    \end{align*}
\end{enumerate}
These generalizations lead to Algorithm~\ref{algo:fairfl-general} below.

\begin{algorithm}[!t]
\caption{General MMD-Fair FedAvg}\label{algo:fairfl-general}
\begin{algorithmic}[1]
\State Initialize $\theta^1$
\State Collect information to estimate $\alpha_{k,j}^a, a\in\mathcal{A}, j\in\mathcal{J}$
\State Communicate $\alpha_{k,j}^a, a\in\mathcal{A}, j\in\mathcal{J}$, to each client $k$
\For{$t = 1, \dots, T$}
\State Construct $\mathcal{Y}_{a,j}^t, a\in\mathcal{A}$, using $\theta^t$
    \State {\color{black}Communicate $\theta^t$ and $\mathcal{Y}_{a,j}^t, a\in\mathcal{A}, j\in\mathcal{J}$, to all clients}
    \State Sample subset $\mathcal{S}\subseteq\{1,\ldots,K\}$ of $S$ clients
    \For{client $k\in\mathcal{S}$ in parallel}
        \State Initialize $\theta_{1,k}^t = \theta^t$
        \For{local iteration $i = 1, \dots, E$}
            \State Compute a stochastic gradient estimator $g_{i,k}^t$ such that:
            \skipnumber{$
            \mathbb{E}\left[g_{i,k}^t| \mathcal{Y}_{a,j}^t,\forall a\in\mathcal{A},\forall j\in\mathcal{J}\right]= \nabla_\theta \big(\mathbb{E}_k[L(h_{ \theta_{i,k}^t}(X),Y)]+ \lambda \sum_{j\in\mathcal{J}}f_k^j( \theta_{i,k}^t;\mathcal{Y}_{0,j}^t,\mathcal{Y}_{1,j}^t)\big)$}
            \State Update $\theta_{i+1,k}^t = \theta_{i,k}^t - \eta_l g_{i,k}^t$
        \EndFor
        \State Communicate $\theta_{E+1,k}^t$ to server
    \EndFor
    \State Update $\theta^{t+1} = \theta^t + \eta_g \left(\frac{1}{S} \sum_{k\in\mathcal{S}} (\theta_{E+1,k}^t - \theta^t)\right)$
\EndFor
\end{algorithmic}
\end{algorithm}
}

\section{Proofs}
We now present all proofs, organized by section.

\subsection{Proofs of Section~\ref{sec:fair-FL}}
\begin{proof}[Proof of Proposition~\ref{proposition:decomposition-impossibility}]
We prove the proposition by constructing a distribution $\mathbb{P}$ for which the equality cannot hold for any dissimilarity measure $\mathcal{D}$ satisfying the identity of indiscernibles and any regularization function $\rho$ with the given properties. For simplicity, assume $K=2$; the extension to higher $K$ is straightforward.  

Assume, for contradiction, that $\mathcal{D}$ satisfies the identity of indiscernibles. Consider the distributions  
\[
\mathbb{P}_1^{h(X) \mid A=0} = \mathbb{P}_2^{h(X) \mid A=1} \neq \mathbb{P}_1^{h(X) \mid A=1} = \mathbb{P}_2^{h(X) \mid A=0}
\]
with mixing weights $\nu_1 = \nu_2 = 0.5$. It is easily verified that  
\[
\frac{1}{2} \mathbb{P}_1^{h(X) \mid A=1} + \frac{1}{2} \mathbb{P}_2^{h(X) \mid A=1} = \frac{1}{2} \mathbb{P}_1^{h(X) \mid A=0} + \frac{1}{2} \mathbb{P}_2^{h(X) \mid A=0}.
\]
Thus,  
\[
\mathcal{D} \left( \frac{1}{2} \mathbb{P}_1^{h(X) \mid A=1} + \frac{1}{2} \mathbb{P}_2^{h(X) \mid A=1}, \frac{1}{2} \mathbb{P}_1^{h(X) \mid A=0} + \frac{1}{2} \mathbb{P}_2^{h(X) \mid A=0} \right) = 0.
\]
However, by the identity of indiscernibles,  
\[
\mathcal{D}(\mathbb{P}_1^{h(X) \mid A=1}, \mathbb{P}_1^{h(X) \mid A=0}) > 0 \quad \text{and} \quad \mathcal{D}(\mathbb{P}_2^{h(X) \mid A=1}, \mathbb{P}_2^{h(X) \mid A=0}) > 0.
\]
Since $\rho(v) = 0$ if and only if $v = 0$, the equality in the proposition statement cannot hold, leading to a contradiction. Thus, the claim follows.
\end{proof}

\subsection{Proofs of Section~\ref{sec:algorithm}}
\begin{proof}[Proof of Lemma~\ref{lemma:mmd-grad}]
An elementary calculation shows that
    \begin{align*}
        &\nabla_\theta f(\theta;\mathcal{Y}_0,\mathcal{Y}_1)\\=&2\nabla_\theta\left(\int_{\mathbb{R}^n} C(h_\theta(x);\mathcal{Y}_0,\mathcal{Y}_1)\mathbb{P}^{X|A=0}(\dd x) - \int_{\mathbb{R}^n} C(h_\theta(x);\mathcal{Y}_0,\mathcal{Y}_1)\mathbb{P}^{X|A=1}(\dd x)\right)\\
        =&2\left(\int_{\mathbb{R}^n} C'(h_\theta(x);\mathcal{Y}_0,\mathcal{Y}_1)\nabla_\theta h_\theta(x)\mathbb{P}^{X|A=0}(\dd x) - \int_{\mathbb{R}^n} C'(h_\theta(x);\mathcal{Y}_0,\mathcal{Y}_1)\nabla_\theta h_\theta(x)\mathbb{P}^{X|A=1}(\dd x)\right)\\
        =&2\Big(\int_{\mathbb{R}^n} \left(\frac{1}{|\mathcal{Y}_0|}\sum_{y\in\mathcal{Y}_0} \kernel^{(1)}(h_\theta(x),y) - \frac{1}{|\mathcal{Y}_1|}\sum_{y\in\mathcal{Y}_1} \kernel^{(1)}(h_\theta(x),y)\right)\nabla_\theta h_\theta(x)\mathbb{P}^{X|A=0}(\dd x) \\& - \int_{\mathbb{R}^n} \left(\frac{1}{|\mathcal{Y}_0|}\sum_{y\in\mathcal{Y}_0} \kernel^{(1)}(h_\theta(x),y) - \frac{1}{|\mathcal{Y}_1|}\sum_{y\in\mathcal{Y}_1} \kernel^{(1)}(h_\theta(x),y)\right)\nabla_\theta h_\theta(x)\mathbb{P}^{X|A=1}(\dd x)\Big),
    \end{align*}
    \paperupdate{where we use $\kernel^{(1)}$ to denote derivative of the kernel with respect to its first argument. Therefore, we find}
    \begin{align}\label{eq:mmdgrad-f}
    \begin{split}
        &\mathbb{E}\left[\nabla_\theta f(\theta;\mathcal{Y}_0,\mathcal{Y}_1)\right]\\
        =&2\Bigg(\int_{\mathbb{R}^n} \Big(\int_{\mathbb{R}^n} \kernel^{(1)}(h_\theta(x),h_\theta(x'))\mathbb{P}^{X|A=0}(\dd x')\\&\hspace{1.5cm}- \int_{\mathbb{R}^n} \kernel^{(1)}(h_\theta(x),h_\theta(x'))\mathbb{P}^{X|A=1}(\dd x')\Big)\nabla_\theta h_\theta(x)\mathbb{P}^{X|A=0}(\dd x) \\& - \int_{\mathbb{R}^n} \Big(\int_{\mathbb{R}^n} \kernel^{(1)}(h_\theta(x),h_\theta(x'))\mathbb{P}^{X|A=0}(\dd x') \\&\hspace{1.5cm}- \int_{\mathbb{R}^n} \kernel^{(1)}(h_\theta(x),h_\theta(x'))\mathbb{P}^{X|A=1}(\dd x')\Big)\nabla_\theta h_\theta(x)\mathbb{P}^{X|A=1}(\dd x)\Bigg),
    \end{split}
    \end{align}
    where we used the identity
    \begin{equation*}
    \mathbb{E}\left[\frac{1}{|\mathcal{Y}_a|}\sum_{y\in\mathcal{Y}_a} \kernel^{(1)}(y',y)\right]=\int_{\mathbb{R}^n} \kernel^{(1)}(y',h_\theta(x))\mathbb{P}^{X|A=a}(\dd x)\quad\forall a\in\mathcal A,\, y'\in\mathbb{R}^n.
    \end{equation*}
    Next, we address the gradient of the fairness loss. By the definition of the MMD metric, we have
\begin{align*}
\nabla_\theta \mathcal{D}_{\rm MMD}(\mathbb{P}^{h_\theta(X)|A=0}, \mathbb{P}^{h_\theta(X)|A=1})^2
=\nabla_\theta &\Big( \int_{\mathbb{R}^n \times \mathbb{R}^n} \kernel(h_\theta(x), h_\theta(x'))\, \mathbb{P}^{X|A=0}(\dd x) \mathbb{P}^{X|A=0}(\dd x') \\
+& \int_{\mathbb{R}^n \times \mathbb{R}^n}  \kernel(h_\theta(x), h_\theta(x')) \, \mathbb{P}^{X|A=1}(\dd x)\mathbb{P}^{X|A=1}(\dd x') \\
    - 2&\int_{\mathbb{R}^n\times \mathbb{R}^n} \kernel(h_\theta(x), h_\theta(x'))\,  \mathbb{P}^{X|A=0}(\dd x) \mathbb{P}^{X|A=1}(\dd x')\Big).
\end{align*}
Interchanging the gradient with the integrals using Leibniz's integral rule, which applies due to the theorem's assumptions, the last expression is equivalent to
\begin{align*}
\nabla_\theta \mathcal{D}_{\rm MMD}(\mathbb{P}^{h_\theta(X)|A=0}, \mathbb{P}^{h_\theta(X)|A=1})^2
=&\Big( \int_{\mathbb{R}^n \times \mathbb{R}^n} \nabla_\theta \kernel(h_\theta(x), h_\theta(x'))\, \mathbb{P}^{X|A=0}(\dd x) \mathbb{P}^{X|A=0}(\dd x') \\
+& \int_{\mathbb{R}^n \times \mathbb{R}^n}  \nabla_\theta \kernel(h_\theta(x), h_\theta(x')) \, \mathbb{P}^{X|A=1}(\dd x)\mathbb{P}^{X|A=1}(\dd x') \\
    - 2&\int_{\mathbb{R}^n\times \mathbb{R}^n} \nabla_\theta \kernel(h_\theta(x), h_\theta(x'))\,  \mathbb{P}^{X|A=0}(\dd x) \mathbb{P}^{X|A=1}(\dd x')\Big)
\end{align*}
By leveraging the chain rule for multivariable functions, we then obtain
\begin{equation*}
\nabla_\theta \kernel\left(h_\theta(x),h_\theta(x')\right)=\kernel^{(1)}\left(h_\theta(x), h_\theta(x')\right)\nabla_\theta h_\theta(x)+\kernel^{(2)}\left(h_\theta(x), h_\theta(x')\right)\nabla_\theta h_\theta(x'),
\end{equation*}
where $\kernel^{(1)}(x,x')=\partial \kernel(x,x')/\partial x $ and $\kernel^{(2)}(x,x')=\partial \kernel(x,x')/\partial x' $. By symmetry of the kernel~$\kernel$, we thus find
\begin{equation*}
\nabla_\theta \kernel\left(h_\theta\left(x\right),h_\theta\left(x'\right)\right)=\kernel^{(1)}\left(h_\theta\left(x\right), h_\theta\left(x'\right)\right)\nabla_\theta h_\theta(x)+\kernel^{(1)}\left(h_\theta\left(x'\right), h_\theta\left(x\right)\right)\nabla_\theta h_\theta(x').
\end{equation*}
In summary, we therefore have
\begin{align*}
&\nabla_\theta \mathcal{D}_{\rm MMD}(\mathbb{P}^{h_\theta(X)|A=0}, \mathbb{P}^{h_\theta(X)|A=1})^2\\
&=\Big( 2\int_{\mathbb{R}^n \times \mathbb{R}^n} \kernel^{(1)}(h_\theta(x), h_\theta(x'))\nabla_\theta h_\theta(x)\, \mathbb{P}^{X|A=0}(\dd x) \mathbb{P}^{X|A=0}(\dd x') \\
&\hspace{1cm}+ 2\int_{\mathbb{R}^n \times \mathbb{R}^n}  \kernel^{(1)}(h_\theta(x), h_\theta(x')) \nabla_\theta h_\theta(x)\, \mathbb{P}^{X|A=1}(\dd x)\mathbb{P}^{X|A=1}(\dd x') \\
    &\hspace{1cm}- 2\int_{\mathbb{R}^n\times \mathbb{R}^n} \kernel^{(1)}(h_\theta(x), h_\theta(x'))\nabla_\theta h_\theta(x)\,  \mathbb{P}^{X|A=0}(\dd x) \mathbb{P}^{X|A=1}(\dd x')\\
    &\hspace{1cm}- 2\int_{\mathbb{R}^n\times \mathbb{R}^n} \kernel^{(1)}(h_\theta(x), h_\theta(x'))\nabla_\theta h_\theta(x)\,  \mathbb{P}^{X|A=1}(\dd x) \mathbb{P}^{X|A=0}(\dd x')\Big).
\end{align*}
The claim then follows by comparing this expression with~\eqref{eq:mmdgrad-f}.
\end{proof}

\subsection{Proofs of Section~\ref{sec:analysis}}
\subsubsection{Proofs of Section~\ref{sec:analysis:convergence}}\label{app:analysis:convergence}
The proof of Theorem~\ref{thm:convergence-fairfl} follows from the following, more general theorem for biased local gradients. It shows the convergence rate of Algorithm~\ref{algo:biased-grads}, which is an abstraction of Algorithm~\ref{algo:fairfl}.
\begin{algorithm}
\caption{FedAvg with Function-Tracking}\label{algo:biased-grads}
\begin{algorithmic}[1]
\State Initialize $\theta^1$
\For{$t = 1, \dots, T$}
    \State Build function estimates $f_k^t$
    \State Communicate $\theta^t$ to clients
    \State Sample subset $\mathcal{S}\subseteq\{1,\ldots, K\}$ of $S$ clients
    \For{$k\in\mathcal{S}$ in parallel}
        \State Initialize $\theta_{1,k}^t = \theta^t$
        \For{epoch $i = 1, \dots, E$}
            \State Compute stochastic gradient $g_k^t(\theta_{i,k}^t)$ of $f_k^t(\theta_{i,k}^t)$
            \State Update $\theta_{i+1,k}^t = \theta_{i,k}^t - \eta_l g_k^t(\theta_{i,k}^t)$
        \EndFor
        \State Communicate $\theta_{E+1,k}^t$ to server
    \EndFor
    \State Update $\theta^{t+1} = \theta^t + \eta_g \left(\frac{1}{S} \sum_{k\in\mathcal{S}} (\theta_{E+1,k}^t - \theta^t)\right)$
\EndFor
\end{algorithmic}
\end{algorithm}

\begin{assumption}[Unbiased Function Estimates]\label{ass:unbiased-function-estimates}
    Let $f_k^t$ be an unbiased estimator for the objective function $F$ in the sense that $\mathbb{E}[\nabla_\theta\sum_{k=1}^Kf_k^t(\theta^t)]=\nabla_\theta F(\theta^t)$, where $\theta^t$ is the $t$-th global iterate of Algorithm~\ref{algo:biased-grads}. Assume further that the variance of $\nabla_\theta\frac{1}{K}\sum_{k=1}^Kf_k^t(\theta^t)$ is bounded by $\omega^2$.
\end{assumption}

\begin{assumption}[G-B Gradient Dissimilarity]\label{ass:gradient-dissimilarity}
    There exist constants $G\geq0$ and $B\geq1$ such that
    \begin{equation*}
    \frac{1}{K}\sum_{k=1}^{K}\|\nabla_\theta f_k^t(\theta)\|^2\leq G^2+B^2 \left\|\nabla_\theta \frac{1}{K}\sum_{k=1}^{K} f_k^t(\theta) \right\|^2.
    \end{equation*}
\end{assumption}

\begin{theorem}[Convergence Rate of Algorithm~\ref{algo:biased-grads}]\label{thm:convergence}
    Assume that the estimators $f_k^t$ and the objective function $F$ are $\beta$-smooth and that $g_k^t$ are unbiased gradient estimators of $f_k^t$ with variance at most $\sigma^2$. 
    If $E\in\mathbb N$, $\eta_l>0$, $\eta_g\geq1$, and the pseudo-stepsize $\Tilde{\eta}=E\eta_l\eta_g$ satsifies $\Tilde{\eta}\leq(60\beta(B^2+1))^{-1}$, and if Assumptions~\ref{ass:unbiased-function-estimates} and~\ref{ass:gradient-dissimilarity} hold, then
    there exist probabilities $\{w_t\}_{t=1}^T$, such that
\begin{equation*}
\mathbb{E}\|\nabla F(\Bar{\theta})\|^2\leq \mathcal{O}\left(\frac{\beta B^2\overline{F}}{T+1}+\frac{\beta M_1\overline{F}^{\frac{1}{2}}}{\sqrt{T+1}}+\frac{\beta^{\frac{2}{3}}M_2\overline{F}^{\frac{2}{3}}}{(T+1)^{\frac{2}{3}}}\right),
\end{equation*}
where $\overline \theta = \theta^t$ with probability~$w_t$,~$\theta^t$ is the $t$-th global iterate of Algorithm~\ref{algo:biased-grads} for every~$t=1,\ldots, T$, while
\begin{align*}
    \overline{F}&=\|\nabla F(\theta^0)\|^2,\\
    M_1^2&=\frac{\sigma^2}{2EK}+ \left(1-\frac{S}{K} \right)\frac{2}{S}G^2+2(B^2+1)\omega^2,\\
    M_2&=\left(\frac{3\sigma^2}{E\eta_g^2}+6(G^2+\omega^2B^2)\right)^{\frac{1}{3}}.
\end{align*}
\end{theorem}
The proof of Theorem~\ref{thm:convergence} is inspired by~\citep{karimireddy2020scaffold}. We give it below for completeness.
\begin{proof}[Proof of Theorem~\ref{thm:convergence}]
    
Before proceeding with the proof, we define some notation. Let $\{\mathcal{F}_t\}_{t\geqslant 0}$ denote the filtration generated by the algorithm. We use the shorthand $\mathbb{E}^t[\cdot] = \mathbb{E}[\cdot \mid \mathcal{F}_t]$ and omit the superscript when clear from context. The objective function is denoted by $F(\theta)$, and at each communication round~$t$, it is approximated by $f^t(\theta) = \frac{1}{K} \sum_{k=1}^{K} f_k^t(\theta)$. The proof follows three steps and parallels the convergence rate proof in~\citep{karimireddy2020scaffold}. To maintain alignment with their approach, we adopt the bound $\Tilde{\eta} \leq (8\beta(B^2+1))^{-1}$ as in~\citep{karimireddy2020scaffold}, which is implied by our theorem assumptions. The bound $\Tilde{\eta} \leq (60\beta(B^2+1))^{-1}$ is used at the final step together with~\cite[Lemma~2]{karimireddy2020scaffold}.

\paragraph{Part 1: Bounding Client Drift}
    In order to quantify how the local models of the clients drift away from the model in the last communication round, we bound
    \begin{equation*}
    \mathcal{E}^t=\mathbb{E}\left[\frac{1}{EK}\sum_{i,k}\|\theta_{i,k}^t-\theta^t\|^2\right].
    \end{equation*}
    Following~\cite{karimireddy2020scaffold}, we can derive the following estimate.
    \begin{align*}
        \mathbb{E}\|\theta_{i,k}^t-\theta^t\|^2=&\mathbb{E}\|\theta_{i-1,k}^t-\theta^t-\eta_l g_k^t(\theta_{i-1,k}^t)\|^2\\
        \leq&\mathbb{E}\|\theta_{i-1,k}^t-\theta^t-\eta_l \nabla f_k^t(\theta_{i-1,k}^t)\|^2+\eta_l^2\sigma^2\\
        \leq&(1+\frac{1}{E-1})\mathbb{E}\|\theta_{i-1,k}^t-\theta^t\|^2 +E\eta_l^2\mathbb{E}\|\nabla f_k^t(\theta_{i-1,k}^t)\|^2+\eta_l^2\sigma^2\\
        =&(1+\frac{1}{E-1})\mathbb{E}\|\theta_{i-1,k}^t-\theta^t\|^2 +\frac{\Tilde{\eta}^2}{E\eta_g}\mathbb{E}\|\nabla f_k^t(\theta_{i-1,k}^t)\|^2+\frac{\Tilde{\eta}^2\sigma^2}{E^2\eta_g^2}\\
        \leq&(1+\frac{1}{E-1})\mathbb{E}\|\theta_{i-1,k}^t-\theta^t\|^2 +\frac{2\Tilde{\eta}^2}{E\eta_g}\mathbb{E}\|\nabla f_k^t(\theta_{i-1,k}^t)-\nabla f_k(\theta^t)\|^2+\frac{2\Tilde{\eta}^2}{E\eta_g}\mathbb{E}\|\nabla f_k^t(\theta^t)\|^2+\frac{\Tilde{\eta}^2\sigma^2}{E^2\eta_g^2}\\
        \leq&(1+\frac{1}{E-1}+\frac{2\Tilde{\eta}^2\beta^2}{E\eta_g})\mathbb{E}\|\theta_{i-1,k}^t-\theta^t\|^2 +\frac{2\Tilde{\eta}^2}{E\eta_g}\mathbb{E}\|\nabla f_k(\theta^t)\|^2+\frac{\Tilde{\eta}^2\sigma^2}{E^2\eta_g^2}\\
        \leq&(1+\frac{2}{E-1})\mathbb{E}\|\theta_{i-1,k}^t-\theta^t\|^2 +\frac{2\Tilde{\eta}^2}{E\eta_g}\mathbb{E}\|\nabla f_k^t(\theta^t)\|^2+\frac{\Tilde{\eta}^2\sigma^2}{E^2\eta_g^2}
    \end{align*}
    We first use the definition of the local update, followed by the mean and variance of the randomness of $g_k^t(\theta_{i-1,k}^t)$. The remaining steps follow from a repeated application of the relaxed triangle inequality~\cite[Lemma~3]{karimireddy2020scaffold}, the definition of $\Tilde{\eta}=E\eta_g\eta_l$, bounds on $\Tilde{\eta}$ and $\eta_g$ and the $\beta$-smoothness of $f_k^t$.
    Unrolling computations, we find that 
    \begin{align*}
        \mathbb{E}\|\theta_{i,k}^t-\theta^t\|^2\leq& \sum_{\tau=1}^{i-1}\left(\frac{2\Tilde{\eta}^2}{E\eta_g}\mathbb{E}\|\nabla f_k^t(\theta^t)\|^2+\frac{\Tilde{\eta}^2\sigma^2}{E^2\eta_g^2}\right)(1+\frac{2}{E-1})^\tau\\
        \leq& \left(\frac{2\Tilde{\eta}^2}{E\eta_g}\mathbb{E}\|\nabla f_k^t(\theta^t)\|^2+\frac{\Tilde{\eta}^2\sigma^2}{E^2\eta_g^2}\right)3E.
    \end{align*}
    Averaging over $i$ and $k$ (recalling again that $\eta_g\geq1$, we get
    \begin{align}
        \mathcal{E}^t\leq&\ \frac{1}{K}\sum_{k\in [K]} 6\Tilde{\eta}^2\mathbb{E}\|\nabla f_k^t(\theta^t)\|^2+\frac{3\Tilde{\eta}^2\sigma^2}{E\eta_g^2}\nonumber\\
        \leq&\frac{3\Tilde{\eta}^2\sigma^2}{E\eta_g^2}+6\Tilde{\eta}^2G^2+6\Tilde{\eta}^2B^2\mathbb{E}\|\nabla f^t(\theta^t)\|^2\nonumber\\
        \leq&\frac{3\Tilde{\eta}^2\sigma^2}{E\eta_g^2}+6\Tilde{\eta}^2(G^2+\omega^2B^2)+6\Tilde{\eta}^2B^2\|\nabla F(\theta^t)\|^2,\label{eq:convergence:client-drift}
    \end{align}
    where the last step uses the bounded mean-variance decomposition together with the bound on variance of $f^t$ and Assumption~\ref{ass:unbiased-function-estimates}.
\paragraph{Part 2: Linearize Progress}
    We next estimate the progress made in each communication round using a linearization of the objective. It satsifies
    \begin{align}
        \langle\nabla F(\theta^t),\mathbb{E}[\theta^{t+1}-\theta^t]\rangle &=-\frac{\Tilde{\eta}}{EK}\sum_{i,k}\langle\nabla F(\theta^t), \mathbb{E}[\nabla f_k^t(\theta_{i,k}^t)]\rangle\nonumber\\
        &\leq -\frac{\Tilde{\eta}}{2}\left\|\nabla F(\theta^t)\right\|^2+\frac{\Tilde{\eta}}{2}\left\|\frac{1}{EK}\sum_{i,k}\mathbb{E}[\nabla f_k^t(\theta_{i,k}^t)]-\nabla F(\theta^t)\right\|^2\nonumber\\
        &=-\frac{\Tilde{\eta}}{2}\left\|\nabla F(\theta^t)\right\|^2+\frac{\Tilde{\eta}}{2}\left\|\frac{1}{EK}\sum_{i,k}\mathbb{E}[\nabla f_k^t(\theta_{i,k}^t)-\nabla f_k^t(\theta^t)+\nabla f_k^t(\theta^t)]-\nabla F(\theta^t)\right\|^2\nonumber\\
        &\leq -\frac{\Tilde{\eta}}{2}\left\|\nabla F(\theta^t)\right\|^2+\Tilde{\eta}\left\|\frac{1}{EK}\sum_{i,k}\mathbb{E}[\nabla f_k^t(\theta_{i,k}^t)-\nabla f_k^t(\theta^t)]\right\|^2+\Tilde{\eta}\left\|\mathbb{E}[\nabla f^t(\theta^t)]-\nabla F(\theta^t)\right\|^2\nonumber\\
        &= -\frac{\Tilde{\eta}}{2}\left\|\nabla F(\theta^t)\right\|^2+\Tilde{\eta}\left\|\frac{1}{EK}\sum_{i,k}\mathbb{E}[\nabla f_k^t(\theta_{i,k}^t)-\nabla f_k^t(\theta^t)]\right\|^2\nonumber\\
        &\leq -\frac{\Tilde{\eta}}{2}\left\|\nabla F(\theta^t)\right\|^2+\frac{\Tilde{\eta}}{EK}\sum_{i,k}\left\|\mathbb{E}[\nabla f_k^t(\theta_{i,k}^t)-\nabla f_k^t(\theta^t)]\right\|^2\nonumber\\
        &\leq -\frac{\Tilde{\eta}}{2}\left\|\nabla F(\theta^t)\right\|^2+\frac{\Tilde{\eta}}{EK}\sum_{i,k}\mathbb{E}\left\|\nabla f_k^t(\theta_{i,k}^t)-\nabla f_k^t(\theta^t)\right\|^2\nonumber\\
        &\leq -\frac{\Tilde{\eta}}{2}\left\|\nabla F(\theta^t)\right\|^2+\Tilde{\eta}\beta^2\mathcal{E}^t.\label{eq:convergence:linear}
    \end{align}
    The first equality follows from the definition of the communication round update, we then use the inequality$-ab\leq\frac{1}{2}((b-a)^2-a^2)$. 
    The third equation follows from Assumption~\ref{ass:unbiased-function-estimates}.
    The inequalities follow from the relaxed triangle inequality~\cite[Lemma~3]{karimireddy2020scaffold} and Jensen's inequality.

\paragraph{Part 3: Second-Order Terms}
    Following again~\cite{karimireddy2020scaffold}
    \begin{align}
    \mathbb{E}[\|\theta^{t+1}-\theta^t\|^2]=&\mathbb{E}[\|-\frac{\Tilde{\eta}}{ES}\sum_{i=1}^E\sum_{k\in \mathcal{S}}g_k^t(\theta_{i,k}^t)\|^2]\nonumber\\
    \leq&\Tilde{\eta}^2\mathbb{E}[\|\frac{1}{ES}\sum_{i=1}^E\sum_{k\in \mathcal{S}}\nabla f_k^t(\theta_{i,k}^t)\|^2] + \frac{\Tilde{\eta}^2\sigma^2}{EK}\nonumber\\
        =&\frac{\Tilde{\eta}^2\sigma^2}{EK}+\Tilde{\eta}^2\mathbb{E}[\|\frac{1}{ES}\sum_{i=1}^E\sum_{k\in \mathcal{S}}\nabla f_k^t(\theta_{i,k}^t)-\nabla f_k^t(\theta^t)+\nabla f_k^t(\theta^t)\|^2]\nonumber\\
        \leq & \frac{\Tilde{\eta}^2\sigma^2}{EK}+2\Tilde{\eta}^2\mathbb{E}[\|\frac{1}{ES}\sum_{i=1}^E\sum_{k\in\mathcal{S}}\nabla f_k^t(\theta_{i,k}^t)-\nabla f_k^t(\theta^t)\|^2]+ 2\Tilde{\eta}^2\mathbb{E}[\|\frac{1}{S}\sum_{k\in\mathcal{S}}\nabla f_k^t(\theta^t)\|^2]\nonumber\\
        \leq & \frac{\Tilde{\eta}^2\sigma^2}{EK}+2\beta^2\Tilde{\eta}^2\mathcal{E}^t+ 2\Tilde{\eta}^2\mathbb{E}[\|\frac{1}{S}\sum_{k\in\mathcal{S}}\nabla f_k^t(\theta^t)\|^2]\nonumber\\
        \leq & \frac{\Tilde{\eta}^2\sigma^2}{EK}+2\beta^2\Tilde{\eta}^2\mathcal{E}^t+ 4\Tilde{\eta}^2\mathbb{E}\|\nabla f^t(\theta^t)\|^2+(1-\frac{S}{K})\frac{4\Tilde{\eta}^2}{SK}\sum_{k=1}^K\mathbb{E}\|\nabla f_k^t(\theta^t)\|^2\nonumber\\
    \leq & \frac{\Tilde{\eta}^2\sigma^2}{EK}+2\Tilde{\eta}^2\beta^2\mathcal{E}^t+4(B^2+1)\Tilde{\eta}^2\mathbb{E}\|\nabla f^t(\theta^t)\|^2+(1-\frac{S}{K})\frac{4\Tilde{\eta}^2}{S}G^2\nonumber\\
    \leq & \frac{\Tilde{\eta}^2\sigma^2}{EK}+2\Tilde{\eta}^2\beta^2\mathcal{E}^t+4(B^2+1)\Tilde{\eta}^2\|\nabla F(\theta^t)\|^2+(1-\frac{S}{K})\frac{4\Tilde{\eta}^2}{S}G^2+4(B^2+1)\Tilde{\eta}^2\omega^2,\label{eq:convergence:second-order}
    \end{align}
    where we first apply the definition of the local updates and the mean and variance of the randomness of $g_k^t(\theta_{i-1,k}^t)$. The second inequality then follows from the relaxed triangle inequality~\cite[Lemma~3]{karimireddy2020scaffold}. We then apply the $\beta$-smoothness of $f_k^t$. The fourth and fifth ineqaulity leverage the mean-variance decomposition and Assumption~\ref{ass:gradient-dissimilarity} respectively. Finally, we apply the bounded mean-variance decomposition together with Assumption~\ref{ass:unbiased-function-estimates}.

\paragraph{Combining all Parts}
We will now combine all parts using the local expansion
\begin{align}
    \mathbb{E}[F(\theta^{t+1})]-F(\theta^t)\leq&\ \langle\nabla F(\theta),\mathbb{E}[\theta^{t+1}-\theta^t]\rangle+\frac{\beta}{2}\mathbb{E}\left[\|\theta^{t+1}-\theta^t\|^2\right]\nonumber\\
    \leq & -\frac{\Tilde{\eta}}{2}\left\|\nabla F(\theta^t)\right\|^2+\Tilde{\eta}\beta^2\mathcal{E}^t\nonumber\\
    &+\frac{\beta}{2}\left(\frac{\Tilde{\eta}^2\sigma^2}{EK}+2\Tilde{\eta}^2\beta^2\mathcal{E}^t+4(B^2+1)\Tilde{\eta}^2\|\nabla F(\theta^t)\|^2+(1-\frac{S}{K})\frac{4\Tilde{\eta}^2}{S}G^2+4(B^2+1)\Tilde{\eta}^2\omega^2\right)\nonumber\\
    =& \left(-\frac{\Tilde{\eta}}{2}+2(B^2+1)\beta\Tilde{\eta}^2\right)\left\|\nabla F(\theta^t)\right\|^2 +\left(\Tilde{\eta}\beta^2+\Tilde{\eta}^2\beta^3\right)\mathcal{E}^t\nonumber\\&+\frac{\Tilde{\eta}^2\beta\sigma^2}{2EK}+(1-\frac{S}{K})\frac{2\Tilde{\eta}^2\beta}{S}G^2+2(B^2+1)\Tilde{\eta}^2\beta\omega^2
    \nonumber\\ \leq&\left(-\frac{\Tilde{\eta}}{2}+2(B^2+1)\beta\Tilde{\eta}^2\right)\left\|\nabla F(\theta^t)\right\|^2 \nonumber\\&+\left(\Tilde{\eta}\beta^2+\Tilde{\eta}^2\beta^3\right)\left(\frac{3\Tilde{\eta}^2\sigma^2}{E\eta_g^2}+6\Tilde{\eta}^2(G^2+\omega^2B^2)+6\Tilde{\eta}^2B^2\|\nabla F(\theta^t)\|^2\right)\nonumber\\&+\frac{\Tilde{\eta}^2\beta\sigma^2}{2EK}+(1-\frac{S}{K})\frac{2\Tilde{\eta}^2\beta}{S}G^2+2(B^2+1)\Tilde{\eta}^2\beta\omega^2\nonumber\\
    =&\left(-\frac{\Tilde{\eta}}{2}+2(B^2+1)\beta\Tilde{\eta}^2+6\Tilde{\eta}^3\beta^2B^2+6\Tilde{\eta}^4\beta^3B^2\right)\left\|\nabla F(\theta^t)\right\|^2 \nonumber\\&+\left(\Tilde{\eta}\beta^2+\Tilde{\eta}^2\beta^3\right)\left(\frac{3\Tilde{\eta}^2\sigma^2}{E\eta_g^2}+6\Tilde{\eta}^2(G^2+\omega^2B^2)\right)\nonumber\\&+\frac{\Tilde{\eta}^2\beta\sigma^2}{2EK}+(1-\frac{S}{K})\frac{2\Tilde{\eta}^2\beta}{S}G^2+2(B^2+1)\Tilde{\eta}^2\beta\omega^2\nonumber\\
    \leq&\ \Tilde{\eta}\left(-\frac{1}{2}+\frac{2}{8}+\frac{6}{64}+\frac{6}{512}\right)\left\|\nabla F(\theta^t)\right\|^2 \nonumber\\&+\Tilde{\eta}\beta^2\left(1+\frac{1}{8}\right)\left(\frac{3\Tilde{\eta}^2\sigma^2}{E\eta_g^2}+6\Tilde{\eta}^2(G^2+\omega^2B^2)\right)\nonumber\\&+\frac{\Tilde{\eta}^2\beta\sigma^2}{2EK}+(1-\frac{S}{K})\frac{2\Tilde{\eta}^2\beta}{S}G^2+2(B^2+1)\Tilde{\eta}^2\beta\omega^2\nonumber\\
    =&\ -\frac{37\Tilde{\eta}}{256}\left\|\nabla F(\theta^t)\right\|^2 \nonumber\\
    &+\frac{9\Tilde{\eta}^3\beta^2}{8}\left(\frac{3\sigma^2}{E\eta_g^2}+6(G^2+\omega^2B^2)\right)\label{eq:convergence:update-bound}\\
    &+\Tilde{\eta}^2\beta\left(\frac{\sigma^2}{2EK}+(1-\frac{S}{K})\frac{2}{S}G^2+2(B^2+1)\omega^2\right).\nonumber
\end{align}
The first inequality follows from the $\beta$-smoothness of $F$. We then apply equations~\eqref{eq:convergence:linear} and~\eqref{eq:convergence:second-order}. The third and fourth inequalities leverage equation~\eqref{eq:convergence:client-drift} and the bound on the pseudo stepsize $\Tilde{\eta}\leq(8\beta(B^2+1))^{-1}$, which implies $\Tilde{\eta}\beta\leq1/8$ and $\Tilde{\eta}\beta B^2\leq1/8$. All equalities are rearranging terms.
Take the average of~\eqref{eq:convergence:update-bound} over all $t=1,\ldots,T$ and apply~\cite[Lemma~2]{karimireddy2020scaffold} with 
\begin{align*}
    \eta&=\frac{37\Tilde{\eta}}{256}\\
    \eta_{\max}&=\frac{37}{2048\beta(B^2+1)}\\
    c_1&\propto\beta\left(\frac{\sigma^2}{2EK}+(1-\frac{S}{K})\frac{2}{S}G^2+2(B^2+1)\omega^2\right)\\
    c_2&\propto\beta^2\left(\frac{3\sigma^2}{E\eta_g^2}+6(G^2+\omega^2B^2)\right).
\end{align*}
We have
$$
\mathbb{E}\|\nabla F(\Bar{\theta}^T)\|^2\leq \mathcal{O}\left(\frac{\beta B^2\overline{F}}{T+1}+\frac{\beta M_1\overline{F}^{\frac{1}{2}}}{\sqrt{T+1}}+\frac{\beta^{\frac{2}{3}}M_2\overline{F}^{\frac{2}{3}}}{(T+1)^{\frac{2}{3}}}\right),
$$
where
\begin{align*}
    \overline{F}&=\|\nabla F(\theta^0)\|^2\\
    M_1^2&=\frac{\sigma^2}{2EK}+(1-\frac{S}{K})\frac{2}{S}G^2+2(B^2+1)\omega^2\\
    M_2&=\left(\frac{3\sigma^2}{E\eta_g^2}+6(G^2+\omega^2B^2)\right)^{\frac{1}{3}}.
\end{align*}
Thus, the claim follows.
\end{proof}
We are now ready to prove Theorem~\ref{thm:convergence-fairfl}.
\begin{proof}[Proof of Theorem~\ref{thm:convergence-fairfl}]
Let $L_L, L_K, L_h$ and $\beta_L, \beta_K, \beta_h$ denote the Lipschnitzness and smoothness parameters of $L$, $\kernel$ and $h_\theta$ respectively.
    We apply Theorem~\ref{thm:convergence}. 
    The unbiasedness in Assumption~\ref{ass:unbiased-function-estimates} is implied by Lemma~\ref{lemma:mmd-grad}. 
    Since $L$, $\kernel$ and $h_\theta$ are all Lipschitz continous, so is $f^t$ with Lipschitz constant $L_LL_h+8\lambda L_KL_h$. This implies that $\|\nabla f^t(\theta)\|\leq L_LL_h+8\lambda L_KL_h$, which by Popoviciu's inequality implies that $\omega^2=\frac{1}{4}(L_LL_h+8\lambda L_KL_h)^2$. The bound on $\|\nabla f^t(\theta)\|$ further implies that Assumption~\ref{ass:gradient-dissimilarity} holds with $G=(L_LL_h+8\lambda L_KL_h)^2$ and $B=1$. The continuity and smoothness also imply that $f_k^t$ and $F$ are $\beta$-smooth for some $\beta$. Finally, we note that $g_{i,k}^t$ in Line 10 of Algorithm~\ref{algo:fairfl} is unbiased and has bounded variance by assumption.
    In summary, the assumptions of Thoerem~\ref{thm:convergence} hold with the specific values
    \begin{align*}
        \omega^2 &=\frac{1}{4}(L_LL_h+8\lambda L_KL_h)^2\\
        \sigma^2 &<\infty\\
        G &= (L_LL_h+8\lambda L_KL_h)^2\\
        L &=1\\
        \beta &=\beta_L L_h + 8\beta_K L_h.
    \end{align*}
\end{proof}

\subsubsection{Proofs of Section~\ref{sec:analysis:privacy}}
\begin{proof}[Proof of Proposition~\ref{proposition:DP-effect}]
The independence of the DP noise and linearity of expectation imply
\begin{align*}
    \mathbb{E}_{{\rm DP}}\left[C(z;\Tilde{\mathcal{Y}}_0,\Tilde{\mathcal{Y}}_1)\right]&=\mathbb{E}_{{\rm DP}}\left[\frac{1}{|\mathcal{Y}_0|}\sum_{y\in\mathcal{Y}_0}\kernel(z,y+\xi_y)-\frac{1}{|\mathcal{Y}_1|}\sum_{y\in\mathcal{Y}_1}\kernel(z,y+\xi_y)\right]\\
    &=\frac{1}{|\mathcal{Y}_0|}\sum_{y\in\mathcal{Y}_0}\mathbb{E}_{{\rm DP}}\left[\kernel(z,y+\xi_y)\right]-\frac{1}{|\mathcal{Y}_1|}\sum_{y\in\mathcal{Y}_1}\mathbb{E}_{{\rm DP}}\left[\kernel(z,y+\xi_y)\right].
\end{align*}
Taking $\Tilde{\kernel}(x,y)=\mathbb{E}_{{\rm DP}}[\kernel(z,y+\xi_y)]$, 
\paperupdate{
we now verify that $\Tilde{\kernel}$ satisfies the conditions of Definition~\ref{def:mmd}.}
We first show that $\Tilde{\kernel}$ is symmetric PSD. For symmetry, note that
\begin{equation*}
\Tilde{\kernel}(x,y)=\mathbb{E}_{{\rm DP}}[\kernel(x,y+\xi)]=\mathbb{E}_{{\rm DP}}[\DiffKernel(x-y-\xi)]=\mathbb{E}_{{\rm DP}}[\DiffKernel(x-y+\xi)]=\mathbb{E}_{{\rm DP}}[\kernel(y,x+\xi)]=\Tilde{\kernel}(y,x),
\end{equation*}
where the middle equality holds because $\mu_{\rm DP}$ is symmetric.
To show that $\Tilde{\kernel}$ is PSD, define the distribution $\mu'$ such that if $a,b\overset{iid.}{\sim}\mu'$, $\xi=a-b\sim\mu_{\rm DP}$.
Observe now that
\begin{align*}
\Tilde{\kernel}(x,y)&=\mathbb{E}^{\xi\sim\mu_{\rm DP}}[\kernel(x,y+\xi)]=\mathbb{E}^{\xi\sim\mu_{\rm DP}}[\DiffKernel(x-y-\xi)]\\&=\mathbb{E}^{a,b\sim\mu'}[\DiffKernel((x+a)-(y+b))]=\mathbb{E}^{a,b\sim\mu'}[\kernel(x+a,y+b)]\\&=\mathbb{E}^{a,b\sim\mu'}[\langle\psi(x+a),\psi(y+b)\rangle_{\mathcal{H}}] \\&= \langle\mathbb{E}^{a\sim\mu'}[\psi(x+a)],\mathbb{E}^{b\sim\mu'}[\psi(y+b)]\rangle_{\mathcal{H}},
\end{align*}
which implies that $\Tilde{\kernel}$ is PSD.
\paperupdate{
It remains to show that there exists a $w\in\mathcal L(\mathbb{R}^n ,[1,\infty))$ such that $\Tilde{\kernel}$ satisfies $\sup_{z\in\mathbb{R}^n} \kernel(z,z')/w(z)<\infty$ for all~$z'\in\mathbb{R}^n$.
Since $\kappa$ is a PSD kernel, we know that 
$$ M=\left(\begin{matrix}\kappa(z,z)&\kappa(z,z')\\\kappa(z,z')&\kappa(z',z')\end{matrix}\right)\succeq 0 \quad\forall z,z'\in\mathbb{R} $$ 
This implies that ${\rm det}(M)\geq0$, which is equivalent to $\kappa(z,z)\kappa(z',z')\geq \kappa(z,z')^2$. Under the assumptions of the Proposition, $\kappa(x,y)=v(x-y)$ for some function $v$, which implies that $v(0)^2\geq\kappa(z,z')^2$, which implies $|\kappa(z,z')|\leq |v(0)|=C<\infty$ for all $z,z'\in\mathbb{R}$. Thus 
$$ \sup_{z\in\mathbb{R}^n}\Tilde{\kappa}(z,z')=\sup_{z\in\mathbb{R}}\int_\mathbb{R}\kappa(z,z'+\xi)\mu_{\rm DP}(\xi){\rm d}\xi\leq \int_\mathbb{R}C\mu_{\rm DP}(\xi){\rm d}\xi\leq C<\infty, $$ 
and the condition $\sup_{w\in\mathbb{R}^n} \tilde{\kappa}(z,z')/w(z)<\infty$ is satisfied with $w(z)=1$. This completes the proof.
}
\end{proof}

\section{Experimentation Details and Further Results}\label{app:hyperparams}
The code for all experiments is available on GitHub~\footnote{\url{https://github.com/yvesrychener/Fair-FL}}. In all experiments, we perform a train-test split on each client, with $25\%$ of the samples used for the test set. All other major details are discussed below.
\subsection{Details for Experiments in Section~\ref{sec:experiments:performance}}
\paragraph{Synthetic Data}
For all methods, we vary the fairness regualrizer from $10^{-5}$ to $100$ in $50$ logarithmically spaced steps. The centralized method uses the SGD optimizer with a stepsize of $0.05$ and trains for $1000$ epochs.
Our method and the \emph{local group fairness} method use the FedAvg Algorithm, with local stepsize $\eta_l=0.05$, global stepsize $\eta_g=1$ and trains for $T=100$ communcation rounds with $E=50$ local epochs each. We do not use client subsampling for our experiments, that is we set $S=K$.

\paragraph{Real Data}
We give the hyperparameters below. Whenever using MMD, we apply a distance-induced kernel, in which case the MMD corresponds to the energy distance.
\begin{itemize}
    \item \textbf{Agnostic}: We use the idea presented by~\cite{mohri2019agnostic} with the algorithm implemented by~\cite{deng2020distributionally}, where we apply local stepsize $\eta_l=0.05$ and train for $T=100$ communcation rounds with $E=10$ local epochs each. At each communication round, the client stepsize $\eta_l$ is multiplied by a factor $0.99$. We do not use client subsampling for our experiments ($S=K$). The local batchsize is set to $100$.
    \item \textbf{Centralized}: We vary the fairness regualrizer from $10^{-5}$ to $10$ in $50$ logarithmically spaced steps. The centralized method uses the SGD optimizer with a stepsize of $0.05$ and trains for $1000$ epochs. We use the full gradient to get the best idea of ``what is possible'' in a centralized setting.
    \item \textbf{Local group fairness}: We vary the fairness regualrizer from $10^{-5}$ to $100$ in $50$ logarithmically spaced steps. The \emph{local group fairness} method uses the FedAvg Algorithm, with local stepsize $\eta_l=0.05$, global stepsize $\eta_g=1$ and trains for $T=100$ communcation rounds with $E=50$ local epochs each. At each communication round, the client stepsize $\eta_l$ is multiplied by a factor $0.99$.  We do not use client subsampling for our experiments ($S=K$). The local batchsize is set to $100$.
    \item \textbf{Ours}: We vary the fairness regualrizer from $10^{-5}$ to $10$ in $50$ logarithmically spaced steps. We use the FedAvg Algorithm, with local stepsize $\eta_l=0.01$, global stepsize $\eta_g=1$ and trains for $T=100$ communcation rounds with $E=50$ local epochs each. At each communication round, the client stepsize $\eta_l$ is multiplied by a factor $0.99$. We do not use client subsampling for our experiments ($S=K$). The local batchsize is set to $100$.
    \item \textbf{FedFB}: We use the code by~\cite{zeng2021improving}. We found a learning rate of 0.001 to perform the best and show results for $\alpha\in\{0.001, 0.05, 0.08, 0.1, 0.2, 0.5, 1, 2\}$. All other hyperparameters take the default values provided by~\cite{zeng2021improving}.\footnote{\url{https://github.com/yzeng58/Improving-Fairness-via-Federated-Learning}} 
    \item \textbf{FairFed} We implement the algorithm Proposed in~\cite{ezzeldin2023fairfed} as closely as possible. In order to have a fair comparison, we choose to use a local MMD regularizer for local fair training. We compare using $\beta\in\{0.5, 1,2\}$ but do not observe a significant effect of varying $\beta$. The local regularization strength~$\lambda$ is varied from $10^{-5}$ to $10$ in $50$ logarithmically spaced steps.
\end{itemize}

\paperupdate{
\subsection{Scatter Plots of Experiments in Section~\ref{sec:experiments:performance}}
While the Pareto frontier is easy to interpet, it hides some valuable information, such as suboptimal performance away from the Pareto frontier. In Figure~\ref{fig:experiments:appendix-scatter} we provide the scatter plot vesion of the plots in Figure~\ref{fig:experiments}.
\begin{figure}
    \centering
    \begin{subfigure}[b]{0.45\textwidth}
    \centering
    \includegraphics[width=0.9\linewidth]{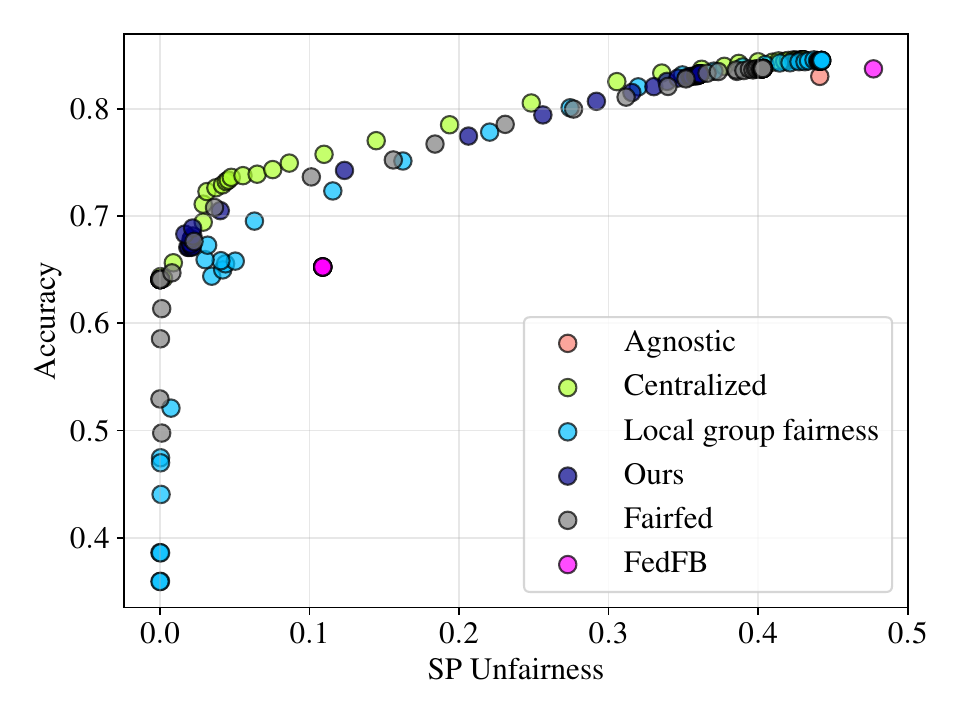}
    \caption{Communities$\&$Crime dataset}
    \end{subfigure}
    \begin{subfigure}[b]{0.45\textwidth}
    \centering
    \includegraphics[width=0.9\linewidth]{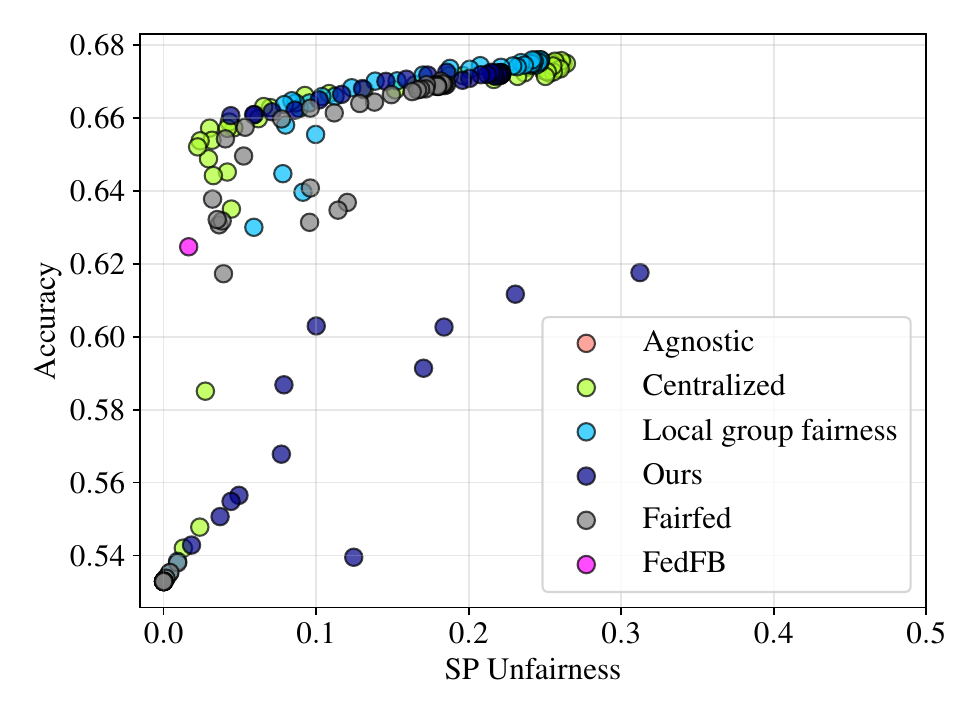}
    \caption{COMPAS dataset}
    \end{subfigure}
    \caption{Performance Comparison on real data}
    \label{fig:experiments:appendix-scatter}
\end{figure}}

\subsection{Ablation Study}\label{app:experiments:ablation}

\begin{figure}
    \centering
    \includegraphics[width=0.5\linewidth]{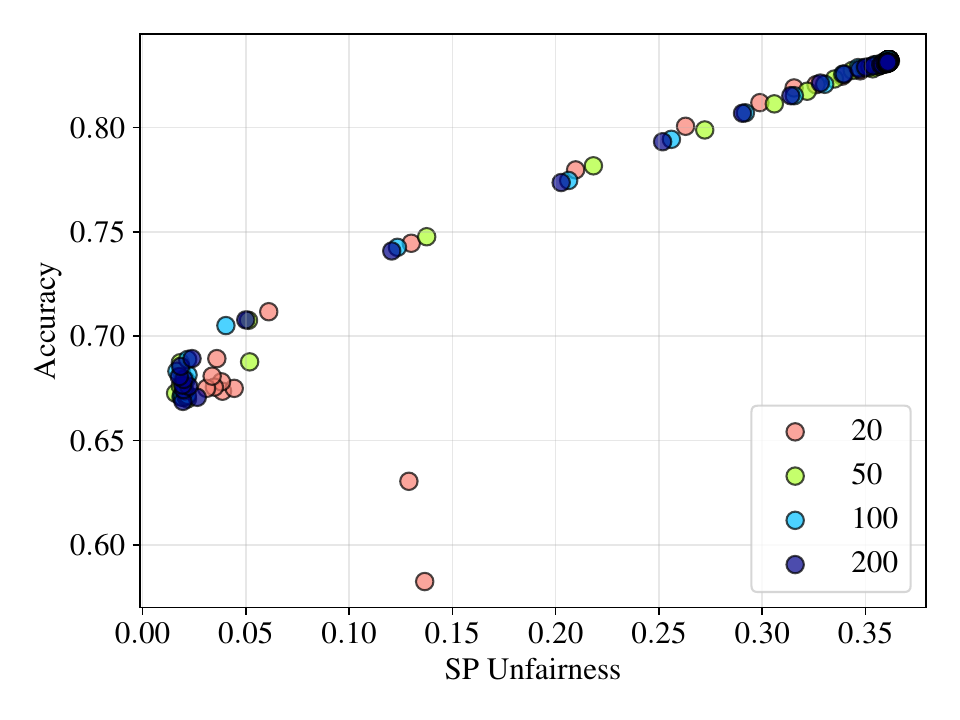}
    \caption{Performance for varying $|\mathcal{Y}_0|$ and $|\mathcal{Y}_1|$}
    \label{fig:ablation:NY}
\end{figure}

\paragraph{Effect of $|\mathcal{Y}_0|$ and $|\mathcal{Y}_1|$}
In Figure~\ref{fig:ablation:NY}, we plot the unfairness-accuracy trade off curve for different sizes of the sets $|\mathcal{Y}_0|$ and $|\mathcal{Y}_1|$. We use the same value for the other hyperparameters as in the experiment on real data. We see that for $20$ samples, the high variance of the gradients results in subpar performance, especially for large values of $\lambda$, which results in low accuracy and relatively high unfairness. However, using $50$ samples is already enough to well estimate the distributions $\mathbb{P}^{h(X)|A=0}$ and  $\mathbb{P}^{h(X)|A=1}$. 
This suggests that good performance can be achieved even with sharing few samples, thus reducing communication cost. This is because $h_\theta(X)$ is a scalar, whose distribution can be well estimated even with few samples available. 

\begin{figure}
    \centering
    \begin{subfigure}[b]{0.5\textwidth}
        \centering
        \includegraphics[height=0.5\linewidth]{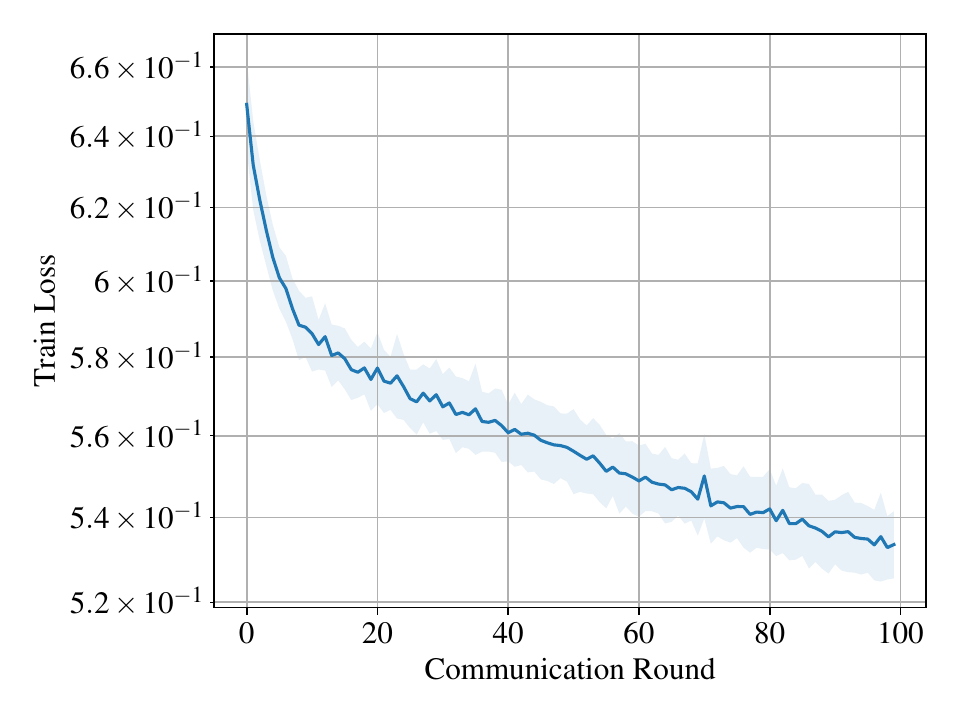}
        \caption{Train Loss}
    \end{subfigure}%
    \begin{subfigure}[b]{0.5\textwidth}
        \centering
        \includegraphics[height=0.5\linewidth]{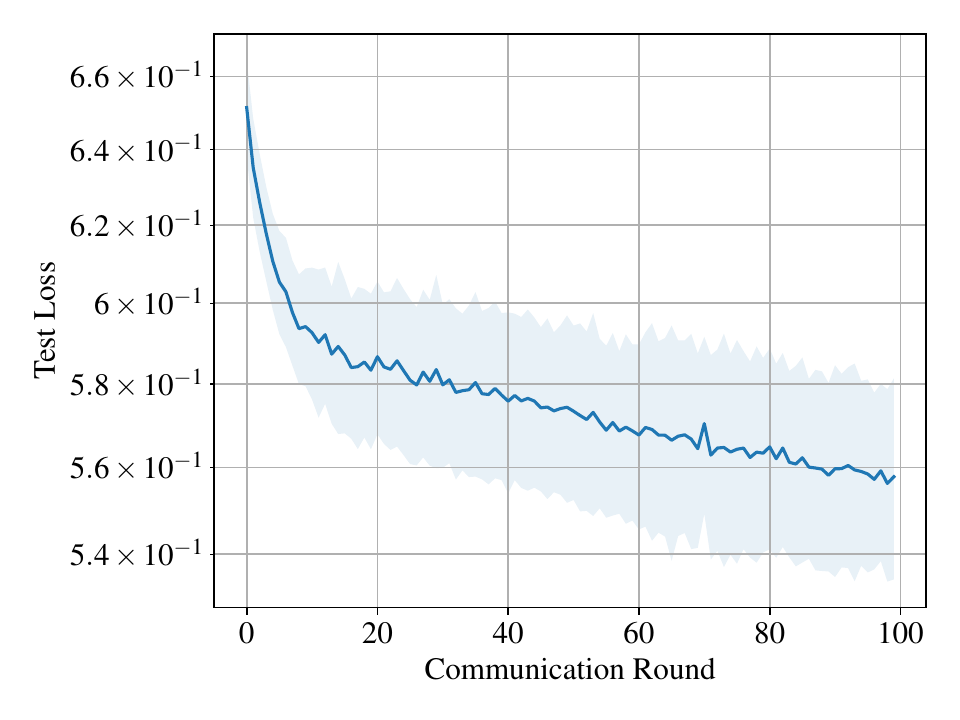}
        \caption{Test Loss}
    \end{subfigure}
    \caption{Train and Test Loss (mean$\pm$standard deviation over 10 runs)}
    \label{fig:convergence}
\end{figure}

\begin{figure}
    \centering
    \begin{subfigure}[b]{0.5\textwidth}
        \centering
        \includegraphics[height=0.5\linewidth]{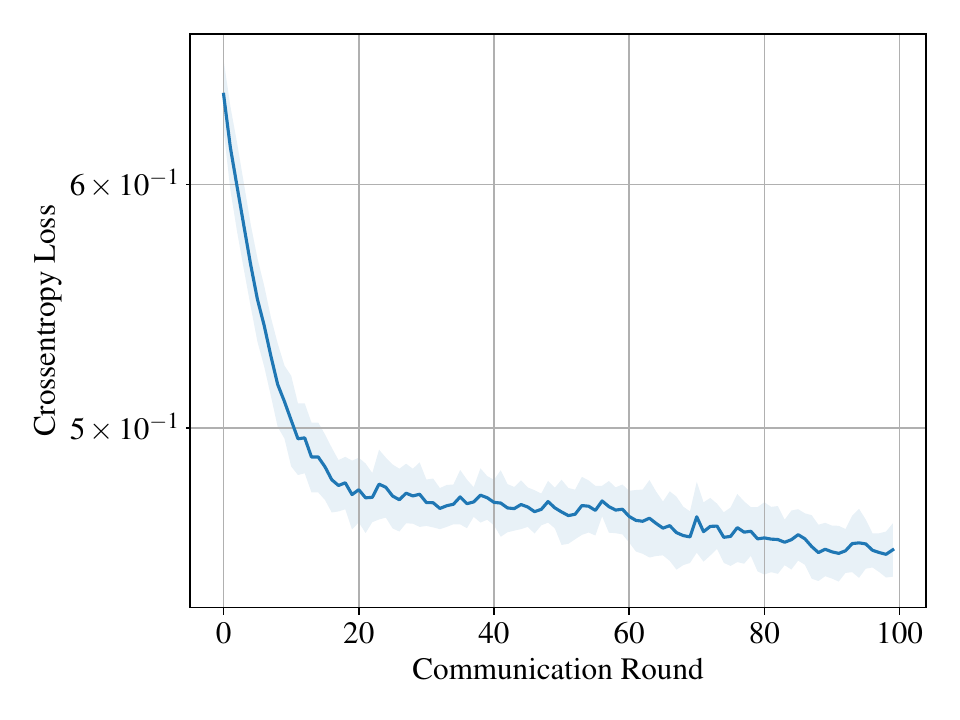}
        \caption{Train Crossentropy}
    \end{subfigure}%
    \begin{subfigure}[b]{0.5\textwidth}
        \centering
        \includegraphics[height=0.5\linewidth]{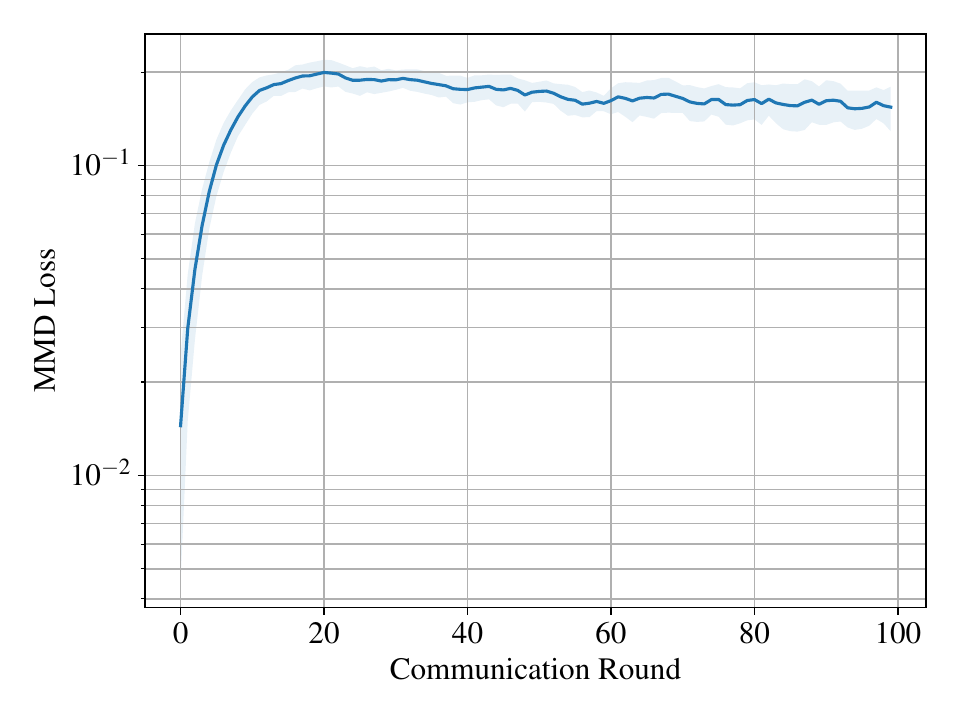}
        \caption{Train MMD}
    \end{subfigure}\\
    \begin{subfigure}[b]{0.5\textwidth}
        \centering
        \includegraphics[height=0.5\linewidth]{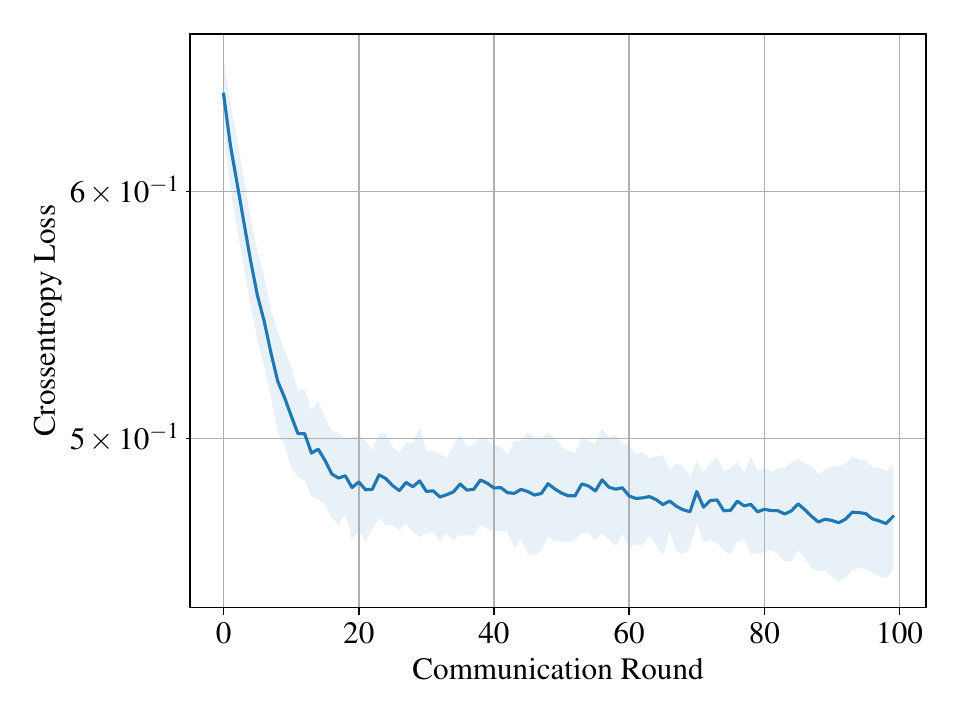}
        \caption{Test Crossentropy}
    \end{subfigure}%
    \begin{subfigure}[b]{0.5\textwidth}
        \centering
        \includegraphics[height=0.5\linewidth]{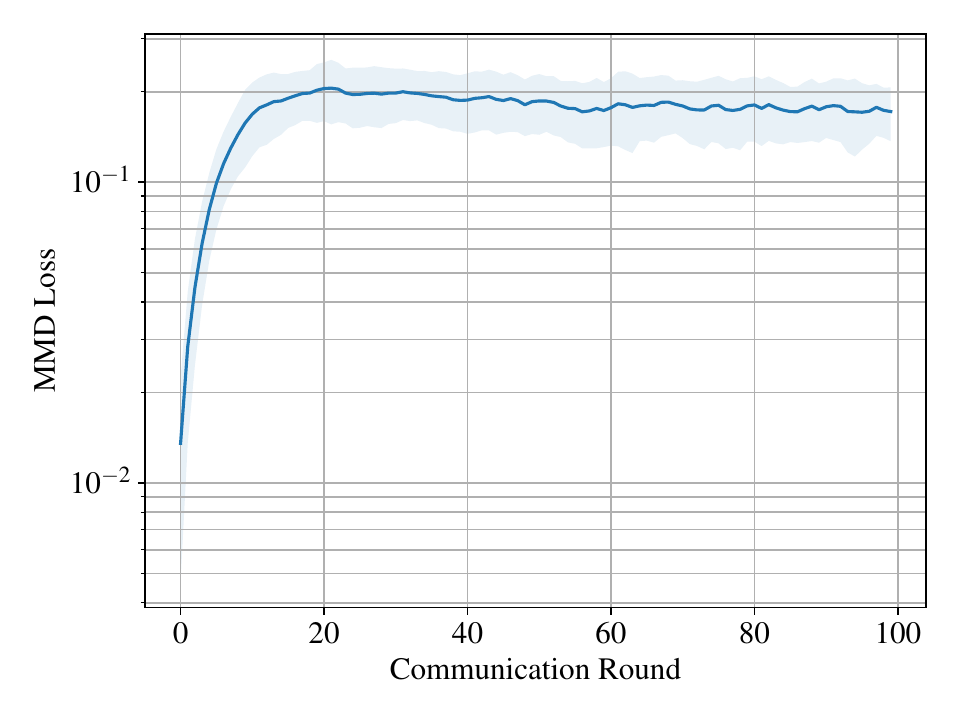}
        \caption{Test MMD}
    \end{subfigure}%
    \caption{Decomposed Train and Test Loss (mean$\pm$standard deviation over 10 runs)}
    \label{fig:appendix:convergence}
\end{figure}

\paragraph{Convergence Rate} To verify the results of Section~\ref{sec:algorithm} and illustrate that Algorithm~\ref{algo:fairfl} minimizes~\eqref{eq:mmd-learning-problem}, we plot the training and testing loss over communication rounds. For this experiment, we again use the Communities$\&$Crime dataset and set $\lambda=0.5$. Figure~\ref{fig:convergence} plots the convergence on the training and testing dataset.  We see that Algorithm~\eqref{algo:fairfl} indeed minimizes the objective of~\eqref{eq:mmd-learning-problem}. Figure~\ref{fig:appendix:convergence} shows more detailed results, where we individually show the crossentropy and MMD part of the loss. 


\paperupdate{
\paragraph{Heterogeneity Ablation} To investigate how the proposed algorithm behaves with increased client heterogeneity, we use two experiments. In the first experiment, we consider a setting with $K=10$ clients with equal weights (that is $\nu_k=\frac{1}{10}$ for all $k=1,\ldots,K$). Further, for all $k=1,\ldots,K$, we assume $\mathbb{P}_k[A=0]=\mathbb{P}_k[A=1]=\frac{1}{2}$ and $X\sim \alpha\mathcal{N}(\mu(k,A),\mathbb{I}) + (1-\alpha)\mathcal{N}(\mu(k,A^c),\mathbb{I})$, where $\mathbb{I}$ is the $10$-dimensional identity matrix, $\mu(k,A)$ is a vector of $1$s ($-1$s) if $k+A$ is even (odd), $A^c$ denotes the complement of $A$ (i.e. $A^c=1$ if $A=0$ and $A^c=0$ otherwise) and $\alpha$ is a heterogeneity parameter: If $\alpha=0.5$, then all clients hold data sampled from the same distribution, if $\alpha=1$ we are in the setting of the synthetic experiment in Figure~\ref{fig:exp:synthetic} and all clients hold different distributions. The label is given by $Y=1$ if $\mathbf{1}^\top X>0$ and $Y=0$ otherwise. We sample 200 datapoints for each client and compare the same methods as in the synthetic experiment. We use $\lambda=50$ and otherwise use the same hyperparameters as in the main text, except that we use $\beta=0.5$ for Fairfed as it shows slightly better performance for low heterogeneity values. The plots are given in Figure~\ref{fig:heterogenity-1}, showing the 95\% confidence interval of mean performance ($mean\pm1.96\ standard\ error$) over 10 runs. We see that our method continues to perform well as client heterogeneity increases, while other baseline methods suffer in performance.

In a second experiment, we use the same data generating process and hyperparameters as for the previous experiment. However, this time, we restrict ourselves to 6 clients with varying dataset size of, in order, 50, 150, 250, 100, 50 and 100 samples. This choice provides a heterogenous dataset size and heterogenous client weights ranging from 0.071 for clients 0 and 4 to 0.357 for client 2, while ensuring that the Bayes classifier is fair. The results are shown in Figure~\ref{fig:heterogenity-2}, showing the 95\% confidence interval of mean performance ($mean\pm1.96\ standard\ error$) over 10 runs. They show that our method continues to perform well if clients contain datasets of different sizes. 

\begin{figure}
    \centering
    \begin{subfigure}[b]{0.45\textwidth}
    \centering
    \includegraphics[width=\linewidth]{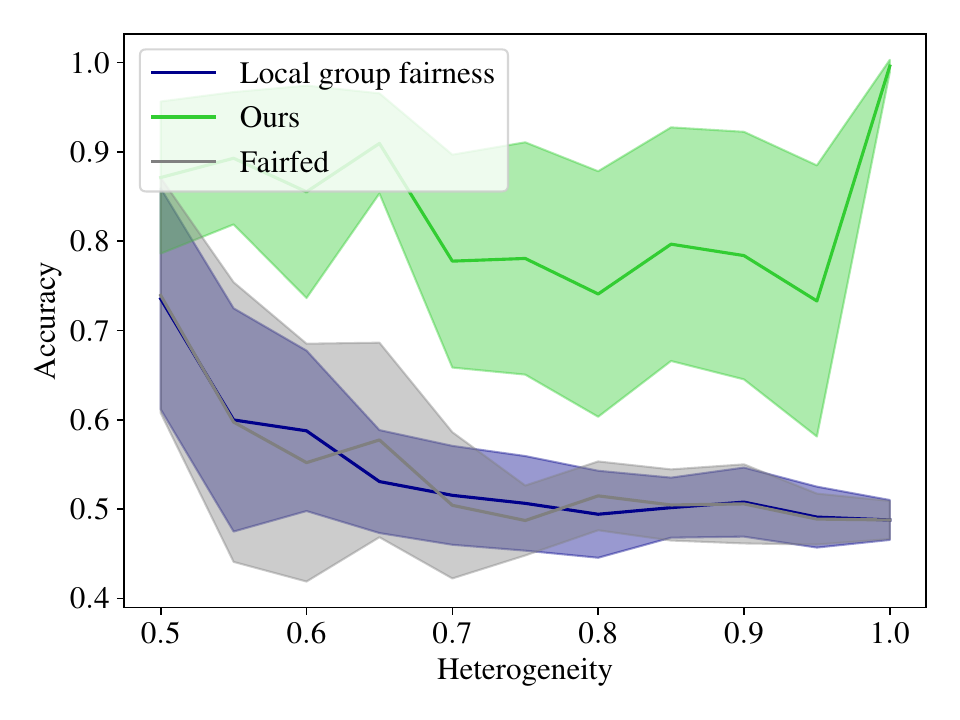}
    \caption{Accuracy}
    \end{subfigure}
    \begin{subfigure}[b]{0.45\textwidth}
    \centering
    \includegraphics[width=\linewidth]{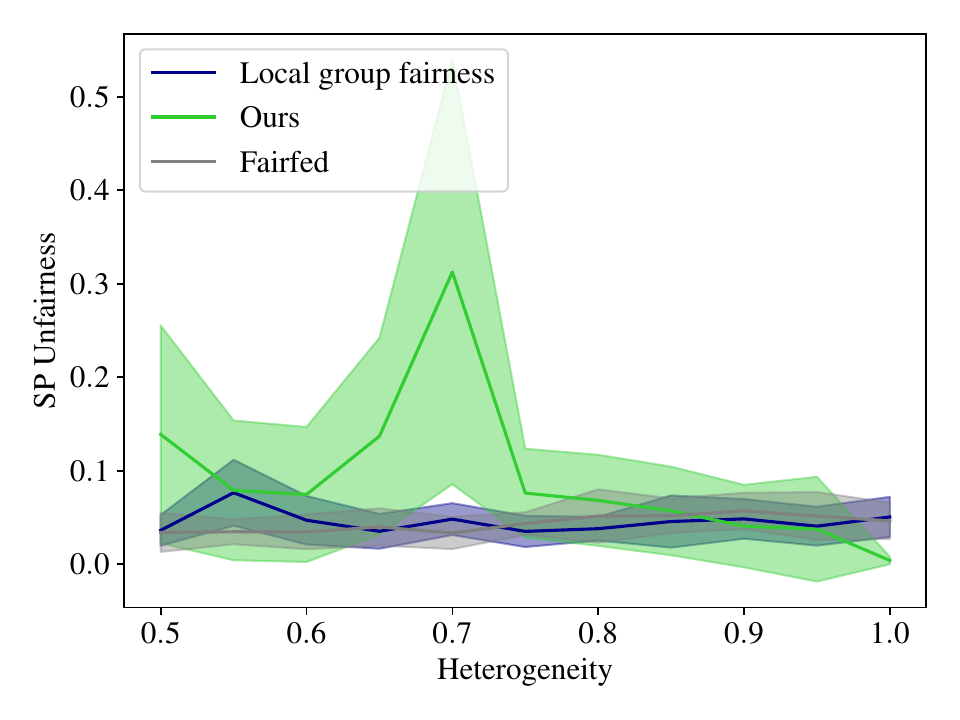}
    \caption{Fairness}
    \end{subfigure}
    \caption{Heterogeneity Ablation for varying Heterogeneity $\alpha$}\label{fig:heterogenity-1}
\end{figure}

\begin{figure}
    \centering
    \begin{subfigure}[b]{0.45\textwidth}
    \centering
    \includegraphics[width=\linewidth]{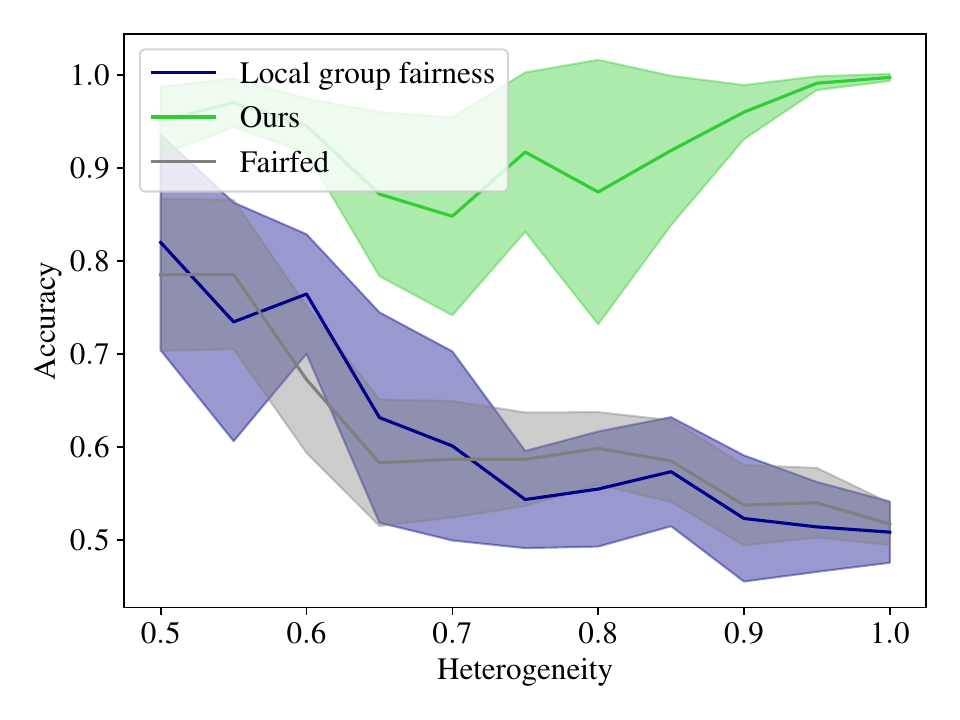}
    \caption{Accuracy}
    \end{subfigure}
    \begin{subfigure}[b]{0.45\textwidth}
    \centering
    \includegraphics[width=\linewidth]{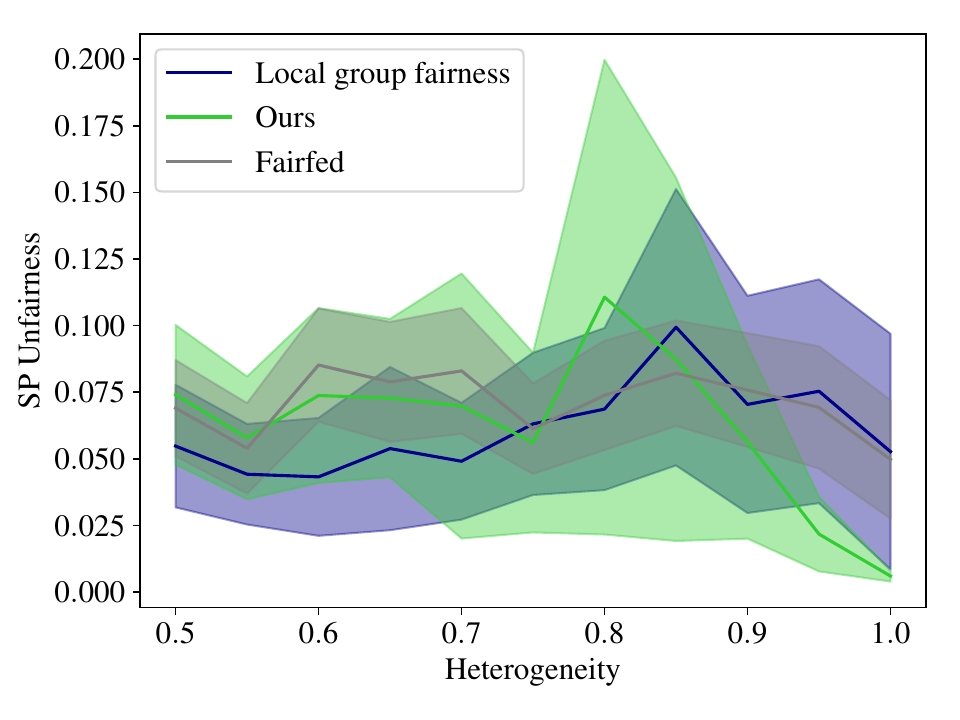}
    \caption{Fairness}
    \end{subfigure}
    \caption{Heterogeneity Ablation for varying Heterogeneity $\alpha$ with differently weighted clients}\label{fig:heterogenity-2}
\end{figure}
}

\end{document}